%% file: main.tex
\documentclass[10pt,twocolumn,letterpaper]{article}

\newcommand\blfootnote[1]{%
  \begingroup
  \renewcommand\thefootnote{}\footnote{#1}%
  \addtocounter{footnote}{-1}%
  \endgroup
}

\usepackage[pagenumbers]{arxi} 

\input{preamble}
\input{macros}

\definecolor{arxiblue}{rgb}{0.21,0.49,0.74}
\usepackage[pagebackref,breaklinks,colorlinks,citecolor=arxiblue]{hyperref}
\usepackage{tabularx}
\usepackage{graphicx}
\usepackage{multirow}
\usepackage{stackengine}
\usepackage{xcolor} 

\title{Text-Conditioned Generative Model of 3D Strand-based Human Hairstyles}

\author{
Vanessa Sklyarova$^{1,2}$ \quad 
Egor Zakharov$^{2}$ \quad 
Otmar Hilliges$^{2}$ \quad 
Michael J. Black$^{1}$ \quad
Justus Thies$^{1,3}$ \quad 
\vspace{0.3cm}\\
$^1$Max Planck Institute for Intelligent Systems \ \ $^2$ETH Zürich \ \ $^3$Technical University of Darmstadt
}

\begin{document}
\input{arxiv_figures/teaser/teaser}
\input{arxiv_parts/abstract}    

\blfootnote{$^\dagger$ \url{https://haar.is.tue.mpg.de/}}

\input{arxiv_parts/intro}

\input{arxiv_parts/related}
\input{arxiv_parts/method}
\input{arxiv_parts/experiments}
\input{arxiv_parts/conclusion}
\input{arxiv_parts/acknowledgements}

{
    \small
    \bibliographystyle{ieeenat_fullname}
    \bibliography{main}
}

\appendix

\vspace{0.5in}
\centerline{\LARGE\bf Supplemental Material}
\vspace{0.1in}

\input{arxiv_parts/method_suppmat}
\input{arxiv_parts/experiments_suppmat}

\end{document}

%% file: preamble.tex
\usepackage{comment}
\usepackage[dvipsnames]{xcolor}

\renewcommand{\paragraph}[1]{\smallskip\noindent\textbf{#1}}

%% file: macros.tex
\newcommand{\EZ}[1]{\iftrue {\color{orange}[EZ: #1]}\else {}\fi} 
\newcommand{\VS}[1]{\iftrue {\color{blue}[VS: #1]}\else {}\fi} 
\newcommand{\JT}[1]{\iftrue {\color{red}[JT: #1]}\else {}\fi} 

\def\model{HAAR\xspace}
\def\modellong{Hair: Automatic Animatable Reconstruction\xspace}

%% file: arxiv_figures/teaser/teaser.tex
\twocolumn[{%
\renewcommand\twocolumn[1][]{#1}%
\maketitle
\begin{center}

\definecolor{mintbg}{rgb}{.63,.79,.95}
\colorlet{lightmintbg}{mintbg!40}

\vspace{-0.6cm}
    \centering
    \captionsetup{type=figure}
        \begin{tabular}{cccc}
           \includegraphics[trim=0 50 0 40,clip,width=.25\textwidth]{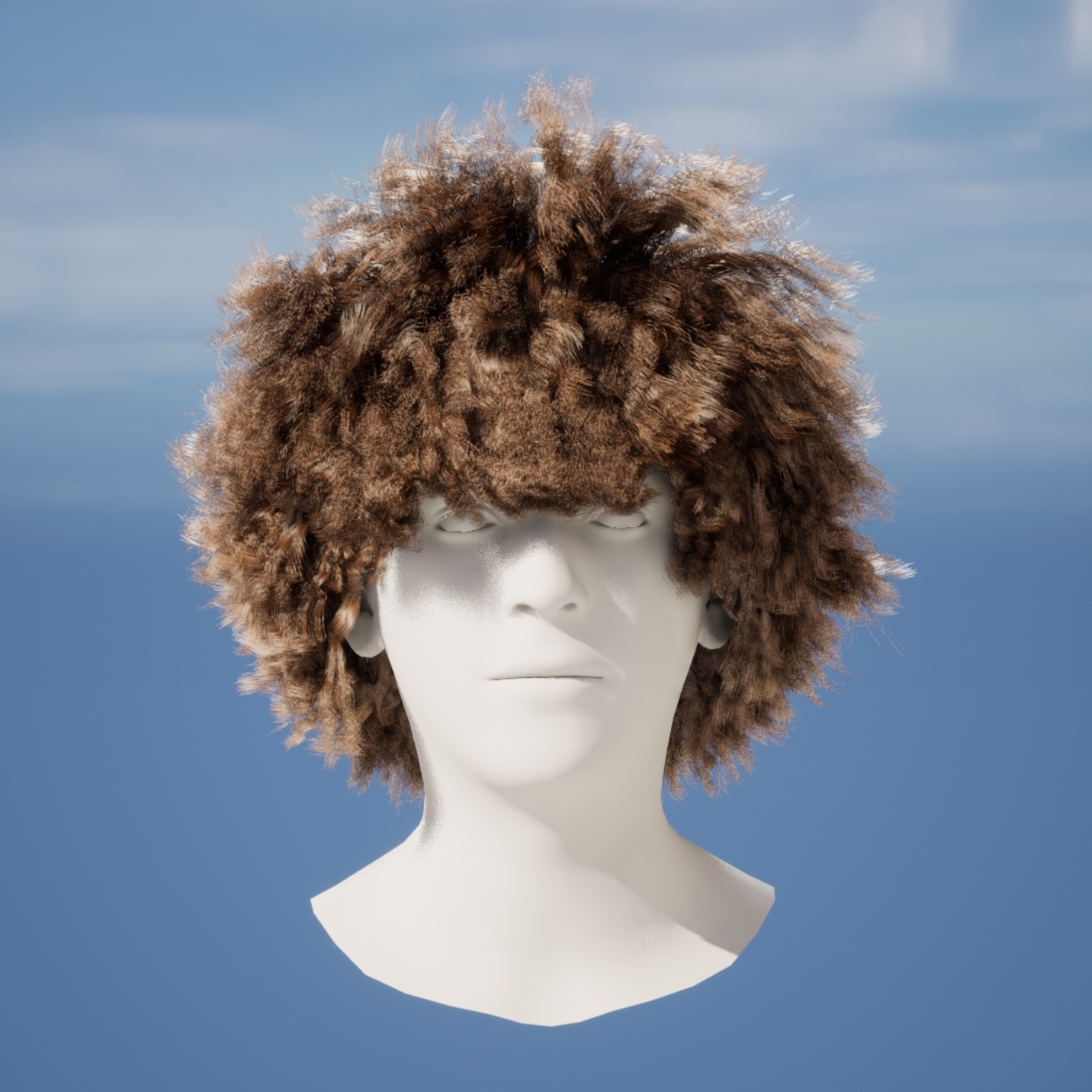}\hspace{-0.43cm}\vspace{-0.13cm} & 
           \includegraphics[trim=0 0 0 90,clip,width=.25\textwidth]{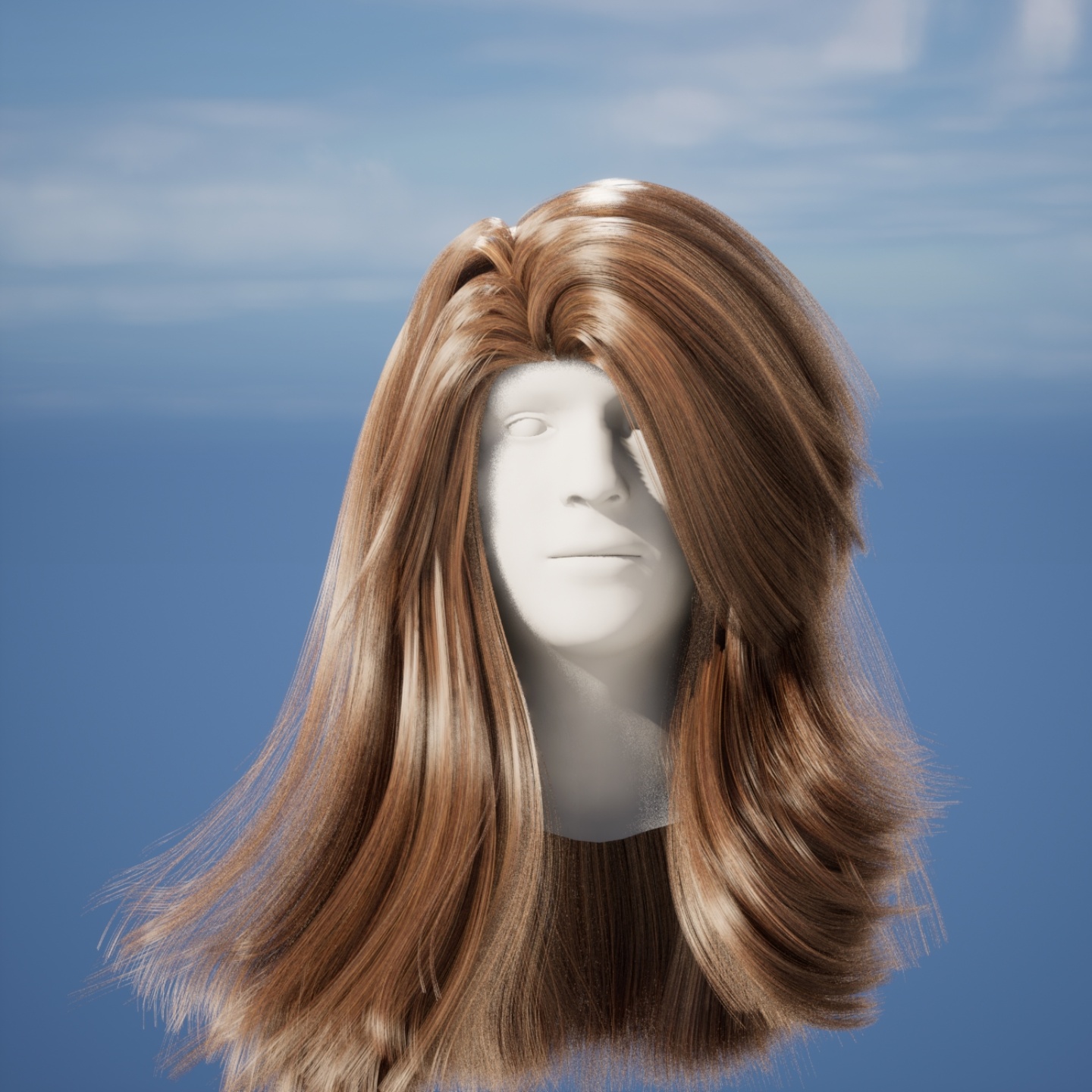}\hspace{-0.43cm} &
           \includegraphics[trim=0 50 0 40,clip,width=.25\textwidth]{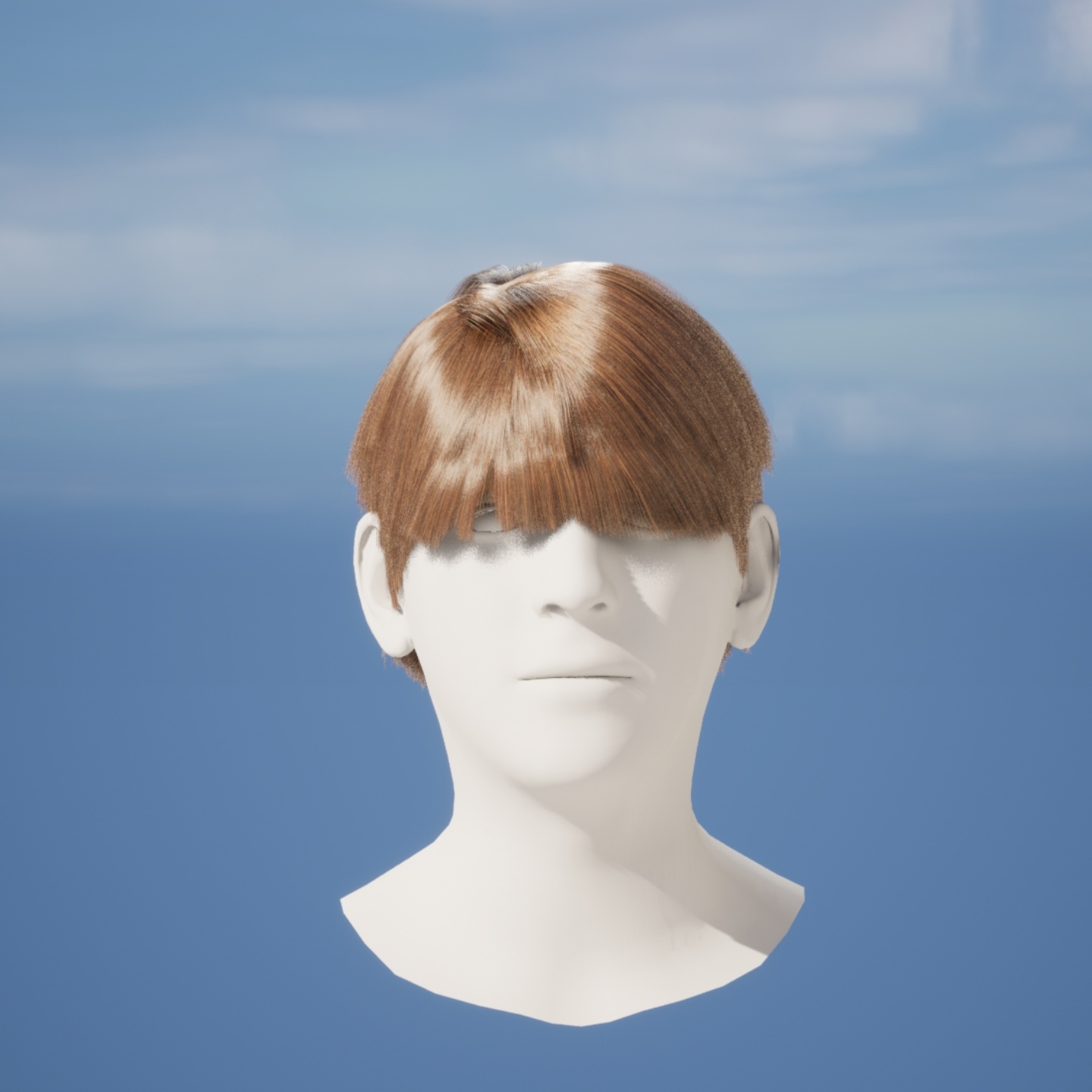}\hspace{-0.43cm} &
           \includegraphics[trim=0 50 0 40,clip,width=.25\textwidth]{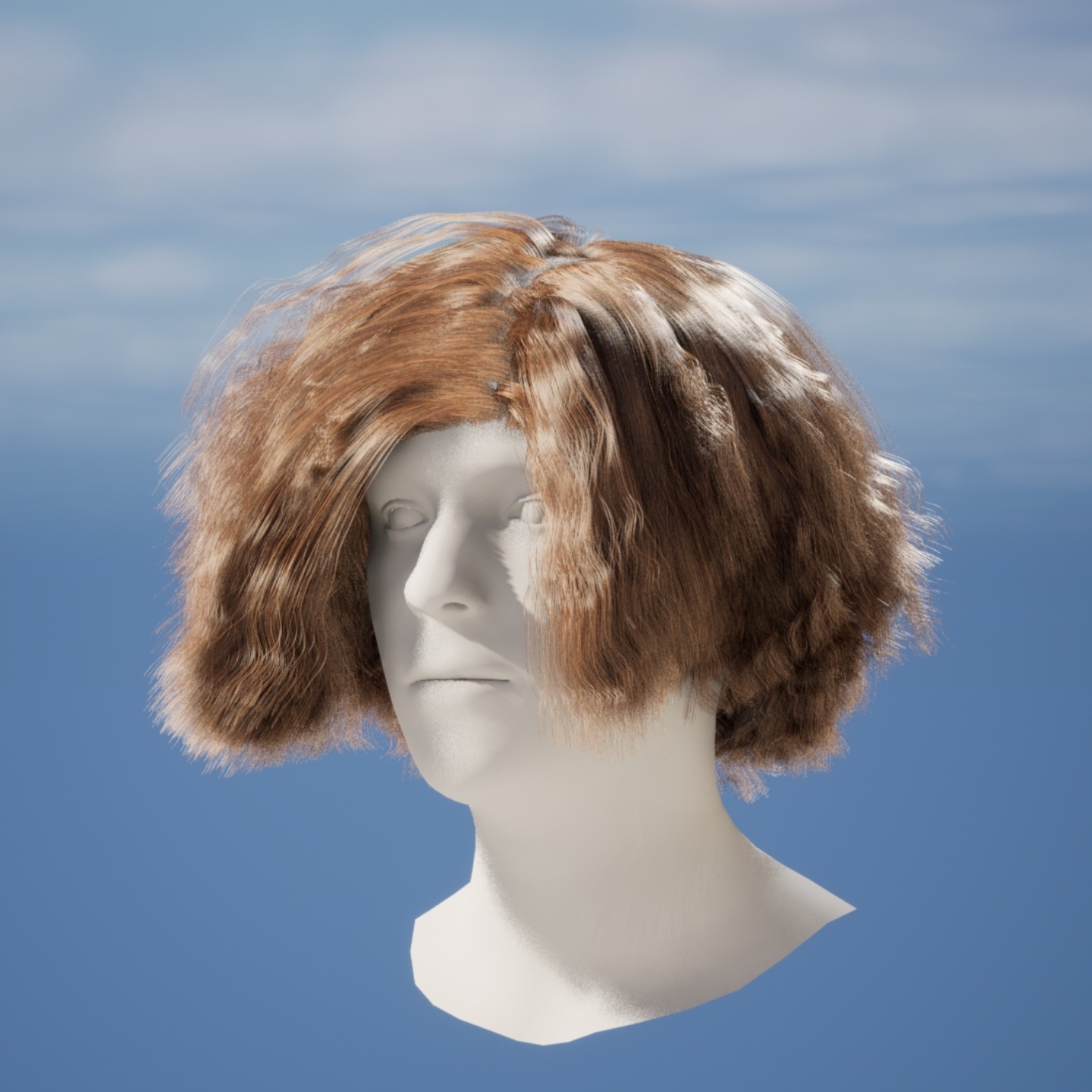}\hspace{-0.43cm} 
           
           \\

            \multicolumn{1}{@{}c}{
                \raisebox{4.05cm}[0pt][0pt]{
                    \setlength{\fboxsep}{2pt}
                        ...afro hairstyle...
                }
                \hspace{-2.23cm}
            } &
            \multicolumn{1}{@{}c}{
                \raisebox{4.05cm}[0pt][0pt]{
                    \setlength{\fboxsep}{2pt}
                        ...voluminous straight hair...\
                }
                \hspace{-0.75cm}
            } &
            \multicolumn{1}{@{}c}{
                \raisebox{4.05cm}[0pt][0pt]{
                    \setlength{\fboxsep}{3pt}
                        ...man haircut...
                }
                \hspace{-2.39cm}
            } &
            \multicolumn{1}{@{}c}{
                \raisebox{4.05cm}[0pt][0pt]{
                    \setlength{\fboxsep}{2pt}
                        ...wavy short hairstyle...
                }
                \hspace{-1.4cm}
            }
            \vspace{-0.42cm}
            
            \\

           \includegraphics[trim=0 50 0 40,clip,width=.25\textwidth]{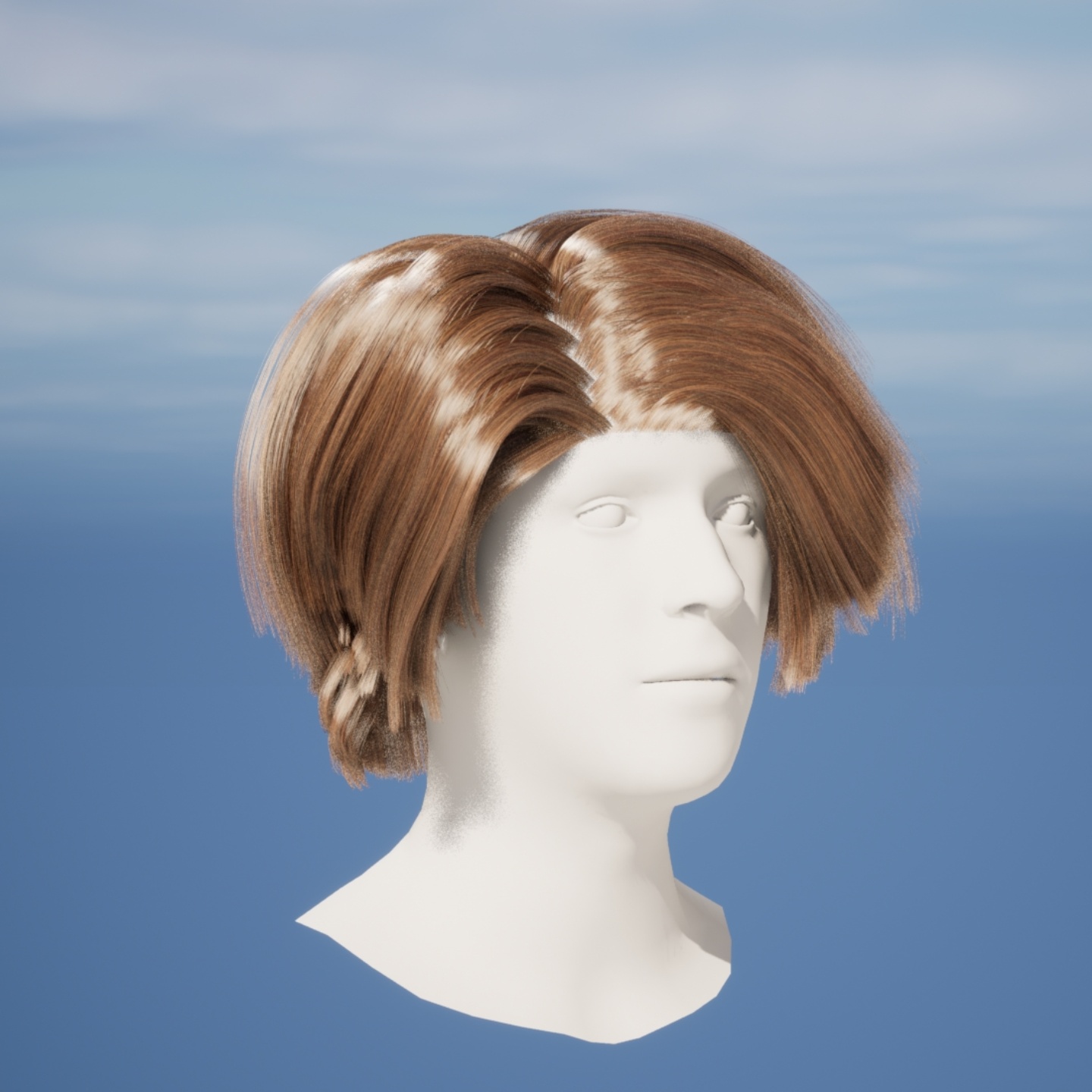}\hspace{-0.43cm}\vspace{-0.13cm} &
           \includegraphics[trim=0 0 0 90,clip,width=.25\textwidth]{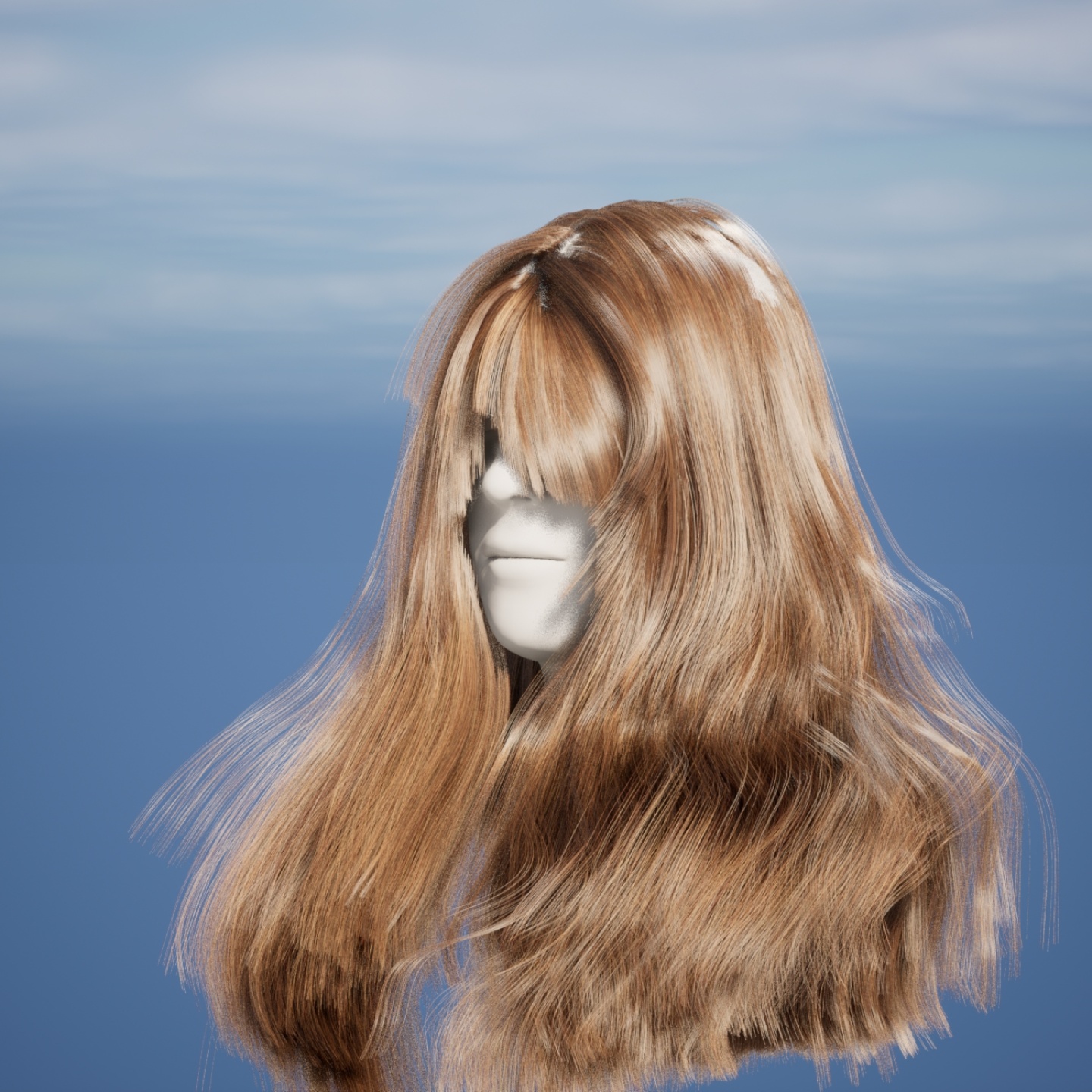}\hspace{-0.43cm} &
           \includegraphics[trim=0 0 0 90,clip,width=.25\textwidth]{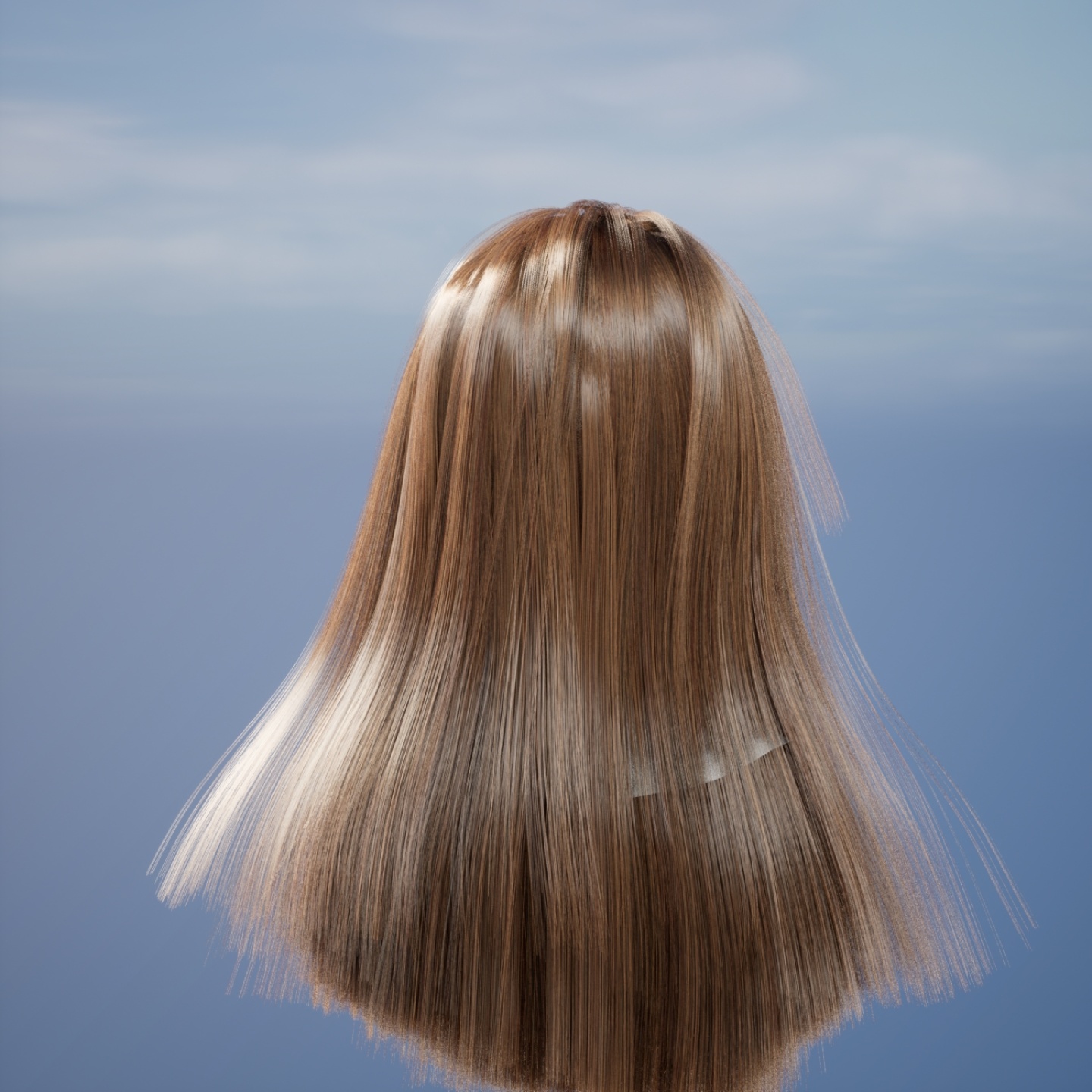}\hspace{-0.43cm} &
           \includegraphics[trim=0 50 0 40,clip,width=.25\textwidth]{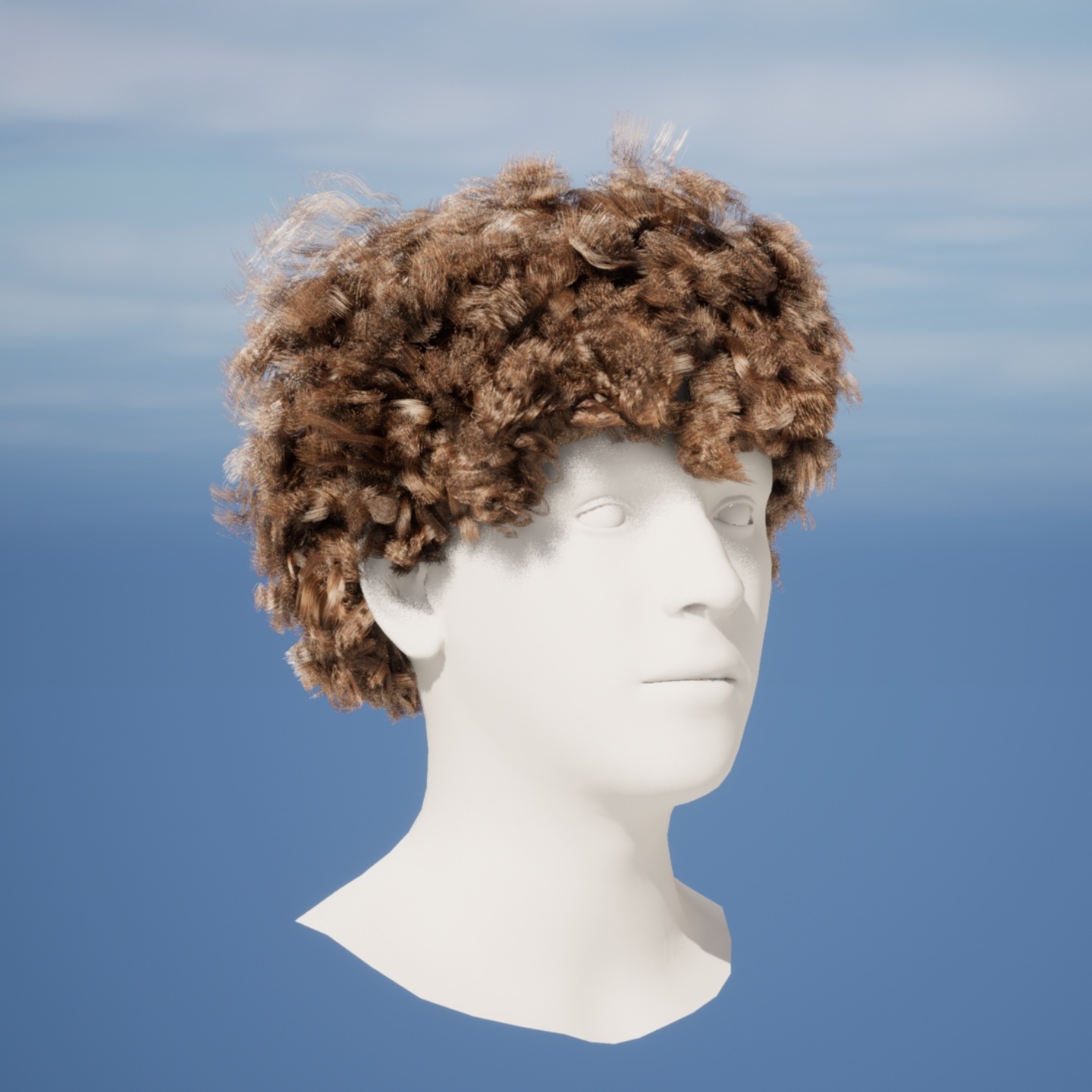}\hspace{-0.43cm}

           \\

            \multicolumn{1}{@{}c}{
                \raisebox{4.05cm}[0pt][0pt]{
                    \setlength{\fboxsep}{3pt}
                        ...bob haircut...
                }
                \hspace{-2.43cm}
            } &
            \multicolumn{1}{@{}c}{
                \raisebox{4.05cm}[0pt][0pt]{
                    \setlength{\fboxsep}{2pt}
                        ...long wavy hairstyle...\
                }
                \hspace{-1.4cm}
            } &
            \multicolumn{1}{@{}c}{
                \raisebox{4.05cm}[0pt][0pt]{
                    \setlength{\fboxsep}{2pt}
                        ...long straight hair...
                }
                \hspace{-1.7cm}
            } &
            \multicolumn{1}{@{}c}{
                \raisebox{4.05cm}[0pt][0pt]{
                    \setlength{\fboxsep}{2pt}
                        ...short curly hairstyle...
                }
                \hspace{-1.42cm}
            }
            \vspace{-0.42cm}
            
            \\
           
    \end{tabular}
    \vspace{-0.2cm}
    \captionof{figure}{Given a text description, our method produces realistic human hairstyles. The usage of a 3D strand-based geometry representation allows it to be easily incorporated into existing computer graphics pipelines for simulation and rendering~\cite{Blender, maya, unrealengine}.}
    \label{fig:teaser}
\end{center}%
}]

%% file: arxiv_parts/abstract.tex
\begin{abstract}
We present HAAR, a new strand-based generative model for 3D human hairstyles. Specifically, based on textual inputs, HAAR produces 3D hairstyles that could be used as production-level assets in modern computer graphics engines. Current AI-based generative models take advantage of powerful 2D priors to reconstruct 3D content in the form of point clouds, meshes, or volumetric functions. However, by using the 2D priors, they are intrinsically limited to only recovering the visual parts. Highly occluded hair structures can not be reconstructed with those methods, and they only model the ``outer shell'', which is not ready to be used in physics-based rendering or simulation pipelines. In contrast, we propose a first text-guided generative method that uses 3D hair strands as an underlying representation. Leveraging 2D visual question-answering (VQA) systems, we automatically annotate synthetic hair models that are generated from a small set of artist-created hairstyles. This allows us to train a latent diffusion model that operates in a common hairstyle UV space. In qualitative and quantitative studies, we demonstrate the capabilities of the proposed model and compare it to existing hairstyle generation approaches. For results, please refer to our project page$^\dagger$.
\end{abstract}
\vspace{-0.8cm}

%% file: arxiv_parts/intro.tex
\section{Introduction}
\label{sec:intro}

There has been rapid progress in creating realistic, animatable 3D face and head avatars from images, video, and text. 
What is still missing is hair.
Existing methods typically represent hair with a coarse mesh geometry, implicit surfaces, or neural radiance fields.
None of these representations are compatible with the strand-based models used by existing rendering systems and do not enable animation of the resulting avatars with natural hair dynamics.
Modeling and generating realistic 3D hair remains a key bottleneck to creating realistic, personalized avatars.
We address this problem with \model (\modellong), which enables the generation of realistic and diverse hairstyles {\em based solely on text descriptions}.
\model is the first text-driven generative model that produces a classical strand-based hair representation that can be immediately imported into rendering systems and animated realistically.
This approach replaces the complex and time-consuming process of manually creating 3D hairstyles with a chat-like text interface that can be used by a novice to create high-quality 3D hair assets.

Previous work exploits generative models as learned priors to create 3D strand-based hair from images, videos, or random noise.
In particular, Neural Haircut~\cite{neuralhaircut} reconstructs high-fidelity hairstyles from smartphone video captures without any specialized equipment by leveraging a pre-trained generative diffusion model.
However, their strand-based generative model does not provide control over the geometry of the resulting hairstyles,  substantially limiting the range of applications.
Recently, GroomGen \cite{GroomGen} introduced an unconditional generative model of hair.
In contrast, we propose the first text-conditioned generative model for strand-based hairstyles that can be used for automated and fast hair asset generation.

Text-conditioned generative models like Stable Diffusion~\cite{stable_diffusion} are widely used for image and video generation and can be used to generate 3D shape from text  \cite{zhang2023teca, lin2023magic3d,qian2023magic123,fantasia3d, chen2023textto3d,tang2023dreamgaussian, tang2023make-it-3d,poole2022dreamfusion,melaskyriazi2023realfusion,liu2023zero1to3,liu2023one2345,huang2024tech} by exploiting Score Distillation Sampling (SDS)~\cite{poole2022dreamfusion}.
These methods convert textual descriptions into 3D assets that, when rendered into multiple views, align with generated 2D images via differentiable rendering. 
These methods represent 3D shapes either as
meshes~\cite{lin2023magic3d,qian2023magic123,fantasia3d}, point clouds~\cite{chen2023textto3d,tang2023dreamgaussian} or volumes~\cite{tang2023make-it-3d,poole2022dreamfusion,melaskyriazi2023realfusion,liu2023zero1to3,liu2023one2345}.
In particular, TECA~\cite{zhang2023teca} demonstrates how hair can be generated from text using a neural radiance field \cite{mildenhall2020nerf}, combined with a traditional mesh-based head model~\cite{flame}.
However, the inherent problem with these SDS-based solutions is that they only capture the outer visible surface of the 3D shape.  
Even volumetric representations do not have a meaningful internal hair structure~\cite{zhang2023teca}. 
Thus, they can not be used for downstream applications like animation in graphics engines~\cite{Blender,maya}.

Instead, what we seek is a solution with the following properties: (1) the hair is represented using classical 3D strands so that the hairstyle is compatible with existing rendering tools, (2) hair is generated from easy-to-use text prompts, (3) the generated hair covers a wide range of diverse and realistic hairstyles, (4) the results are more realistic than current generative models based SDS.
To this end, we develop a text-guided generation method that produces strand-based hairstyles via a latent diffusion model.
Specifically, we devise a latent diffusion model following the unconditional model used in Neural Haircut~\cite{neuralhaircut}.
A hairstyle is represented on the scalp of a 3D head model as a texture map where the values of the texture map correspond to the latent representation of 3D hair strands.
The individual strands are defined in a latent space of a VAE that captures the geometric variation in the hair strand shape.
To generate novel hair texture maps, we infer a diffusion network that takes a noise input and text conditioning.
From the generated hair texture map, we can sample individual latent strands and reconstruct the corresponding 3D hair strands.
There are three remaining, interrelated, problems to address: 
(1) We need a dataset of 3D hairstyles to train the VAE and diffusion model. 
(2) We need training data of hairstyles with text descriptions to relate hairstyles to our representation. 
(3) We need a method to condition generated hair on text. 
We address each of these problems.
First, we combine three different 3D hair datasets and augment the data to construct a training set of about 10K 3D hairstyles.
Second, one of our key novelties is in how we obtain hairstyle descriptions.
Here, we leverage a large vision-language model (VLM) \cite{liu2023llava} to generate hairstyle descriptions from images rendered from the 3D dataset.
Unfortunately, existing visual question-answering (VQA) systems~\cite{liu2023llava, liu2023improvedllava, li2022blip} are inaccurate and do not produce coherent hairstyle descriptions.
To address these problems, we design a custom data-annotation pipeline that uses a pre-generated set of prompts that we feed into a VQA system~\cite{liu2023improvedllava} and produce final annotations by combining their responses in a single textual description.
Finally, we train a diffusion model to produce the hair texture encoding conditioned on the encoding of textual hairstyle descriptions.

As Figure~\ref{fig:teaser} illustrates, our strand-based representation can be used in classical computer graphics pipelines to realistically densify and render the hair~\cite{Blender, maya, unrealengine}.
We also show how the latent representation of hair can be leveraged to perform various semantic manipulations, such as up-sampling the number of strands in the generated hairstyle (resulting in better quality than the classical graphics methods) or editing hairstyles with text prompts. 
We perform quantitative comparisons with Neural Haircut as well as an ablation study to understand which design choices are critical.
In contrast to SDS-based methods like TECA, \model is significantly more efficient, requiring seconds instead of hours to generate the hairstyle. 

\medskip
\noindent
Our contributions can be summarized as follows:
\begin{itemize}
    \item We propose a first text-conditioned diffusion model for realistic 3D strand-based hairstyle generation,
    \item We showcase how the learned latent hairstyle representations can be used for semantic editing,
    \item We developed a method for accurate and automated annotation of synthetic hairstyle assets using off-the-shelf VQA systems.
\end{itemize}
The model will be available for research purposes.

%% file: arxiv_parts/related.tex
\section{Related work}

Recently, multiple text-to-3D approaches~\cite{zhang2023teca, lin2023magic3d,qian2023magic123,fantasia3d, chen2023textto3d,tang2023dreamgaussian, tang2023make-it-3d,poole2022dreamfusion,melaskyriazi2023realfusion,liu2023zero1to3,liu2023one2345,huang2024tech} have emerged that were inspired by the success of text-guided image generation~\cite{stable_diffusion,dalle,clip,imagen}.
A body of work of particular interest to us is the one that uses image-space guidance to generate 3D shapes in a learning-by-synthesis paradigm.
Initially, these methods used CLIP~\cite{clip} embeddings shared between images and text to ensure that the results generated by the model adhere to the textual description~\cite{text2mesh,aneja2022clipface,hong2022avatarclip}.
However, the Score Distillation Sampling procedure (SDS)~\cite{poole2022dreamfusion} has recently gained more popularity since it could leverage text-to-image generative diffusion models, such as Stable Diffusion~\cite{stable_diffusion}, to guide the creation of 3D assets from text, achieving higher quality.
Multiple concurrent methods employ this SDS approach to map textual description into a human avatar~\cite{huang2024tech, liao2023tada, zhang2023teca, kolotouros2023dreamhuman, cao2023dreamavatar}.
In particular, the TECA~\cite{zhang2023teca} system focuses on generating volumetric hairstyles in the form of neural radiance fields (NeRFs)~\cite{mildenhall2020nerf}.
However, these approaches can only generate the outer visible surface of the hair without internal structure, which prevents it from being used out-of-the-box in downstream applications, such as simulation and physics-based rendering.
Moreover, the SDS procedure used to produce the reconstructions is notoriously slow and may require hours of optimization to achieve convergence for a given textual prompt.
Our approach is significantly more efficient, and is capable of generating and realistically rendering the hairstyles given textual prompts in less than a minute.

In contrast to the methods mentioned above, we also generate the hairstyles in the form of strands.
Strand-accurate hair modeling has manifold applications in computer vision and graphics as it allows subsequent physics-based rendering and simulation using off-the-shelf tools~\cite{Blender, maya, unrealengine}.
One of the primary use cases for the strand-based generative modeling has historically been the 3D hair reconstruction systems~\cite{neuralhaircut, Zhou2018HairNetSH, Wu2022NeuralHDHairAH, Yang2019DynamicHM, Saito20183DHS, Hu2015SingleviewHM, Nam2019StrandAccurateMH, neuralstrands, Kuang2022DeepMVSHairDH, zheng2023hairstep, shen2023CT2Hair}.
Among the settings where it is most often used is the so-called one-shot case, where a hairstyle must be predicted using only a single image~\cite{Hu2015SingleviewHM, Wu2022NeuralHDHairAH, zheng2023hairstep}.
Approaches that tackle it leverage synthetic datasets of strand-based assets to train the models and then employ detailed cues extracted from the images, such as orientation maps~\cite{Paris2004CaptureOH}, to guide the generation process. 
However, these systems are unsuitable for semantics-based or even unconditional generation of hairstyles, as they rely heavily on these cues for guidance.
A group of methods that is more closely related to ours is Neural Haircut~\cite{neuralhaircut} and GroomGen~\cite{GroomGen}, in which a synthetic dataset of hairstyle assets is leveraged to train an unconditional generative model~\cite{stable_diffusion, Karras2022ElucidatingTD, kingma2022autoencoding}.
While useful for regularizing multi-view hair reconstruction~\cite{neuralhaircut}, the degree of control over the synthesized output in such methods is missing.
Our work addresses the issue of controllability in generative models for hair and is the first one to provide strand-based hairstyle generation capabilities given textual descriptions.

%% file: arxiv_parts/method.tex
\section{Method}

\input{arxiv_figures/schemes/scheme}

Given a textual description that contains information about hair curliness, length, and style, our method generates realistic strand-based hair assets.
The resulting hairstyles can be immediately used in computer graphics tools that can render and animate the hair in a physically plausible fashion.
Our pipeline is depicted in \Cref{fig:scheme}.
At its core is a latent diffusion model, which is conditioned on a hairstyle text embedding.
It operates on a latent space that is constructed via a Variational Autoencoder (VAE)~\cite{kingma2022autoencoding}.
Following~\cite{neuralstrands}, this VAE is trained to embed the geometry of individual strands into a lower-dimensional latent space.
During inference, the diffusion model generates this representation from Gaussian noise and the input text prompt, which is then upsampled to increase the number of strands and decoded using a VAE decoder to retrieve the 3D hair strands.

\subsection{Hairstyle parametrization.}
We represent a 3D hairstyle as a set of 3D hair strands that are uniformly distributed over the scalp.
Specifically, we define a hair map $H$ with resolution $256~\times~256$ that corresponds to a scalp region of the 3D head model.
Within this map, each pixel stores a single hair strand $S$ as a polyline. 
As mentioned previously, our diffusion model is not directly operating on these 3D 
polylines, but on their compressed latent embeddings $z$.
To produce $z$ that encodes the strand $S$, we first convert the latter into the local basis defined by the Frenet frame of the face where the strand root is located.
On this normalized data, we train a variational auto-encoder, which gives us access to an encoder $\mathcal{E}(S)$ and a decoder $\mathcal{G}(z)$.
Using the encoder $\mathcal{E}(S)$, we encode the individual hair strands in the hair map $H$, resulting in a latent map $Z$ that has the same spatial resolution. 
The decoded strand-based hair map is then denoted as $\hat{H}$.
In summary, with a slight abuse of notation, the maps are related to each other as follows: $Z = \mathcal{E}(H)$, and $\hat{H} = \mathcal{G}(Z)$.

\subsection{Conditional Hair Diffusion Model}

We use a pre-trained text encoder $\tau$~\cite{li2022blip}, that encodes the hairstyle description $P$ into the embedding $\tau (P)$.
This embedding is used as conditioning to the denoising network via a cross-attention mechanism:
\begin{equation}
    \text{Attention}(Q, K, V) = \text{softmax}\, \left( \frac{QK^{T}}{\sqrt{d}} \right) \cdot V,
\end{equation}
where $Q=W_{Q}^{(i)} \cdot \phi_{i} (Z_{t})$, $K=W_{K}^{(i)}\cdot \tau(P)$, $V=W_{V}^{(i)}\cdot \tau (P)$ with learnable projection matrices $W_{Q}^{(i)}, W_{K}^{(i)}, W_{V}^{(i)}$.
The denoising network is a 2D U-Net~\cite{pix2pix2017}, where $\phi_{i} (Z_t)$ denotes $i$-th intermediate representations of the U-Net produced for the latent hair map $Z_t$ at the denoising step $t$.
%
%
For our training, we employ the EDM~\cite{Karras2022ElucidatingTD} formulation, following~\cite{neuralhaircut}.
We denote the latent hair map with noise as $Z_t = Z + \epsilon \cdot \sigma_t$, where $\epsilon \sim \mathcal{N}(0, I)$, and $\sigma_t$ is the noise strength.
We then use a denoiser $\mathcal{D}$ to predict the output:
\begin{equation}
    \mathcal{D}_\theta (Z_t, \sigma_t, P) = c^s_t \cdot Z_t + c^o_t \cdot \mathcal{F}_\theta \big( c^i_t \cdot Z_t, c^n_t, \tau (P) \big),
\end{equation}
where the $c^s_t$, $c^o_t$, $c^i_t$ and $c^n_t$ are the preconditioning factors for the noise level $\sigma_t$ that follow~\cite{Karras2022ElucidatingTD}, and $\mathcal{F}_\theta$ denotes a U-Net network.
The optimization problem is defined as:
\begin{equation}
    \min_\theta\ \mathbb{E}_{\sigma_t, \epsilon, Z, P} \big[ \lambda_t \cdot \| \mathcal{D}_\theta (Z_t, \sigma_t, P) - Z \|_2^2 \big],
\end{equation}
where $\lambda_t$ denotes a weighting factor for a given noise level.

\subsection{Upsampling}
Due to the limited amount of available 3D hairstyles, the diffusion model is trained on a downsampled latent hair map $Z'$ with resolution $32\times 32$ and, thus, only generates so-called 'guiding hair strands'.
To increase the number of strands in the generated results, we upsample the latent hair map to the resolution of $512\times 512$.
A common way of upsampling a strand-based hairstyle to increase the number of strands is via interpolation between individual polylines.

In modern computer graphics engines~\cite{Blender, maya} multiple approaches, such as Nearest Neighbour (NN) and bilinear interpolation are used.
Applying these interpolation schemes leads to over-smoothing or clumping results.
In some more advanced pipelines, these schemes are combined with distance measures based on the proximity of strand origins or the similarity of the curves.
Additionally, Blender and Maya~\cite{Blender, maya} introduce an option of adding noise into the interpolation results to further prevent clumping of the hair strands and increase realism. 
However, the described interpolation procedure requires a lot of manual effort and needs to be done for each hairstyle separately to obtain optimal parameters and resolve undesired penetrations.

In this work, we propose an automatic approach with interpolation of the hairstyle in \textit{latent space} by blending between nearest neighbor and bilinear interpolation schemes.
In this way, we aim to preserve the local structure of strands near a partition and apply smoothing in regions with similar strand directions.
To calculate the blending weights, we first compute the cosine similarity between neighboring 3D hair strands on the mesh grid and apply the non-linear function $f(\cdot)$ to control the influence of the particular interpolation type, which we empirically derived to be as follows:
\begin{equation}
    f(x) = \begin{cases}
    1-1.63\cdot x^{5} &\text{where $x \le 0.9$} \\
    0.4-0.4\cdot x & x > 0.9,
    \end{cases}
\end{equation}
where $x$ is the cosine similarity.
Our final interpolation for each point on the mesh grid is defined as a blending between the nearest neighbor and bilinear interpolations with the weight $f(x)$ and $(1-f(x))$ correspondingly.
The defined upsampling method ensures that in the vicinity of a partition, the weight of the nearest neighbor decreases linearly, and then diminishes at a polynomial rate.
As a result of this scheme, we obtain realistic geometry in the regions with low similarity among strands.
On top of that, we add Gaussian noise to the interpolated latents to increase the hair strands diversity, resulting in a more natural look.

\subsection{Data generation}

\input{arxiv_figures/schemes/dataset_scheme}

\paragraph{3D hairstyle data.}
For training and evaluating the diffusion model, we use a small artist-created hairstyle dataset, that consists of 40 high-quality hairstyles with around 100,000 strands. 
To increase the diversity, we combine it with two publicly available datasets: CT2Hair~\cite{shen2023CT2Hair} and USC-HairSalon~\cite{Hu2015SingleviewHM} that consist of 10 and 343 hairstyles, respectively.
We align the three datasets to the same parametric head model and additionally augment each hairstyle using realistic squeezing, stretching, cutting, and curliness augmentations.
In total, we train the model on 9825 hairstyles. 

\paragraph{Hairstyle description.}
As these 3D hairstyles do not come with textual annotations, we use the VQA model LLaVA~\cite{liu2023llava, liu2023improvedllava} to automatically produce hairstyle descriptions from a set of predefined questions (see Figure~\ref{fig:dataset_scheme}).
To do that, we first render all collected hairstyles using Blender~\cite{Blender} from frontal and back camera views.
We use the standard head model and neutral shading for hairstyles to prevent bias to any particular type of hairstyle because of color or gender information.
With the help of ChatGPT~\cite{chatgpt}, we design a set of prompts, that include specific questions about length, texture, hairstyle type, bang, etc., as well as a set of general questions about historical meaning, professional look, occasions for such hairstyle, celebrities with similar type to increase generalization and variability of our conditioning model.
We then use a random subset of these prompts for each hairstyle in the dataset to increase the diversity of annotations.
For a full list of prompts that were used, please refer to the suppl. material.
The quality of visual systems is highly restricted by the diversity of data used during training. 
We have observed in our experiments that the accuracy of the produced hair captions is relatively low, or they contain very broad descriptions.
In particular, we have noticed that the existing VQA systems have problems accurately reasoning about the hair length or the side of the parting.
To improve the quality of VQA answers, similarly to~\cite{Zhu2023ChatGPTAB}, we add an additional 
system prompt \textit{``If you are not sure say it honestly. Do not imagine any contents that are not in the image''}, which decreases the likelihood of the model hallucinating its responses.
Further, we have observed that the VQA system works better when it does not use information from the previous answers.
That allows us to not accumulate erroneous descriptions during the annotation session.
We have also observed that the LLAVA model is biased toward confirming the provided descriptions instead of reasoning, so introducing a set of choices to the prompts substantially improves the results.
Finally, we calculate the embeddings of the resulting hairstyle descriptions $P$ using a BLIP encoder $\tau$ for both frontal and back views and average them to produce the conditioning used during training.

\subsection{Training details}
To train the diffusion model, we sample a batch of hairstyles at each iteration, align them on a mesh grid of $256\times256$ resolution, and, then, subsample it into a size of $32\times32$. 
By training the diffusion model on these subsampled hairstyles we improve convergence and avoid overfitting. 
To accelerate the training, we use the soft Min-SNR~\cite{Min-SNR}  weighting strategy. It tackles the conflicting directions in optimization by using an adaptive loss weighting strategy. For more details, please refer to the original Min-SNR paper~\cite{Min-SNR}. 
%
To evaluate the performance, we utilize an Exponential Moving Average (EMA) model and Euler Ancestral Sampling with $50$ steps. 
The whole method is trained for about 5 days on a single NVIDIA A100, which corresponds to 160,000 iterations.
Additional details are in the suppl. material.

%% file: arxiv_figures/schemes/scheme.tex
\begin{figure*}[t!]
    \centering
    \includegraphics[,clip,width=\linewidth]{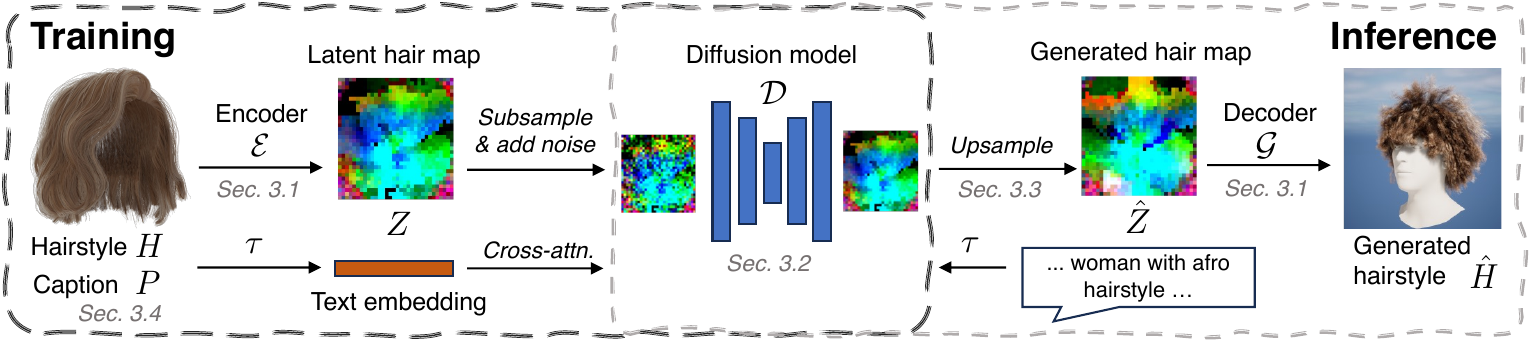}
    \vspace{-0.4cm}
    \caption{
    \textbf{Overview.}
    We present our new method for text-guided and strand-based hair generation. For each hairstyle $H$ in the training set, we produce latent hair maps $Z$ and annotate them with textual captions $P$ using off-the-shelf VQA systems~\cite{liu2023improvedllava} and our custom annotation pipeline. 
    Then, we train a conditional diffusion model $\mathcal{D}$~\cite{Karras2022ElucidatingTD} to generate the \emph{guiding strands} in this latent space and use a latent upsampling procedure to reconstruct dense hairstyles that contain up to a hundred thousand strands given textual descriptions.
    The generated hairstyles are then rendered using off-the-shelf computer graphics techniques~\cite{unrealengine}.
    }
    \label{fig:scheme}
\end{figure*}

%% file: arxiv_figures/schemes/dataset_scheme.tex
\begin{figure}
    \begin{center}
    \includegraphics[trim={0cm 8.2cm 16.75cm 0cm},clip,width=\linewidth]{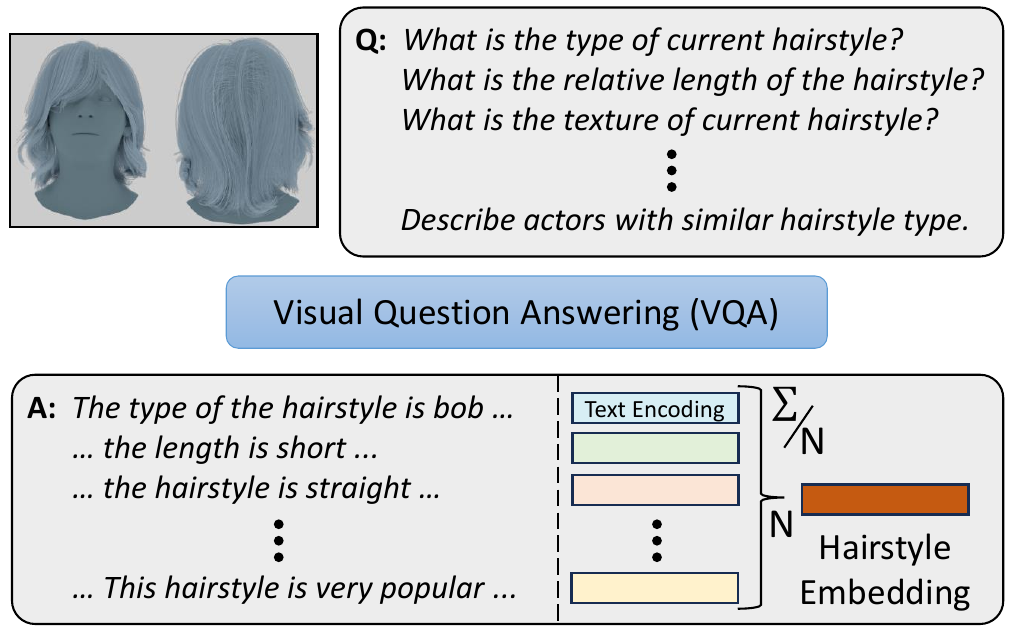}
    \vspace{-0.4cm}
    \caption{\textbf{Dataset collection.} Rendered from frontal and back view hairstyles along with a predefined set of questions Q are sent through VQA~\cite{liu2023llava, liu2023improvedllava} to obtain hairstyle description, which is further encoded using frozen text encoder network~\cite{li2022blip}.} 
    \label{fig:dataset_scheme}
    \end{center}
\end{figure}

%% file: arxiv_parts/experiments.tex
\section{Experiments}

\input{arxiv_figures/comparison/comparison_main}
\subsection{Evaluation}

We compare our method against competing approaches for generative hair modeling: TECA~\cite{zhang2023teca} and Neural Haircut~\cite{neuralhaircut}. 
TECA creates a compositional avatar that includes separate geometries for hair, body, and cloth using only a text description.
This method represents hair using neural radiance fields (NeRF)~\cite{mildenhall2020nerf} and focuses on the visual quality of generated avatars, not geometry reconstruction.
Moreover, it takes multiple hours to generate a single sample using TECA because they rely on Score Distillation Sampling~\cite{poole2022dreamfusion}.
In our case, we concentrate on physically plausible geometry for the hair and require around 4.3 seconds to generate a hairstyle.
Neural Haircut focuses on the reconstruction of realistic 3D hairstyles with a strand-based representation using monocular video or multi-view images captured under unconstrained lighting conditions. 
In this work, authors exploit a diffusion model to obtain some prior knowledge for better reconstruction quality.
In contrast to our approach, the quality of the diffusion model is limited by the amount of data, the size of the model architecture, and the chosen training strategy.
This model is unconditional, and thus cannot control the generated hairstyles.

\paragraph{Quality of unconditional diffusion.} 
To compare the quality of the unconditional diffusion model, we re-train Neural Haircut~\cite{neuralhaircut} on the same training data and with the same scalp parametrization as our method.
We evaluate the distance of the generated hairstyles to the training distribution using Minimum Matching Distance (MMD)~\cite{mmd} as well as coverage (Cov)~\cite{mmd} metrics.
We use the 1-Nearest Neighbor accuracy (1-NNA)~\cite{NNA} metric, which is a leave-one-out accuracy of the 1-NN classifier that assesses if two provided distributions are identical. The best quality is achieved for values closer to 0.5.
Suppose, we have two datasets of generated and reference hairstyles denoted as $S_{g}$ and $S_{r}$, where $| S_{g} | = | S_{r} |$.
Then, the described metrics are defined as: 
\begin{equation}
    \footnotesize
    \text{MMD}(S_{g}, S_{r}) = \frac{1}{| S_{r} | } \sum_{y \in S_{r}} \min_{x \in S_{g}} D(x, y)
\end{equation}
\begin{equation}
    \footnotesize
    \text{COV}(S_{g}, S_{r}) = \frac{1}{| S_{r} |} | 
    \{ \arg\min_{y \in S_{r}} D(x, y) | x \in S_{g} \}|
\end{equation}
\begin{equation}
    \footnotesize
1-\text{NNA}(S_{g}, S_{r})= \frac{\sum_{x \in S_{g}}\mathbb{I}[N_{x}\in S_{g}]+\sum_{y \in S_{r}}\mathbb{I}[N_{y}\in S_{r}]}{\mid S_{g}\mid +\mid S_{r}\mid},
\end{equation}
where $\mathbb{I}(\cdot)$ is an indicator function, $N_{F}$ is the nearest neighbor in set $S_{r}\cup S_{g} \setminus {F}$ and $D$ is the squared distance between distributions, computed in the latent space of the VAE.
\input{arxiv_figures/tab_uncond_comp}
In \Cref{tab:uncond_comp}, we show the comparison based on these metrics. Our method generates samples closer to the ground-truth distribution with higher diversity.
\input{arxiv_figures/upsample/upsample}

Finally, we conducted a user study. Participants were presented 40 randomly sampled hairstyle pairs obtained using Neural Haircut~\cite{neuralhaircut} and our method. We collected more than 1,200 responses on the question \textit{``Which hairstyle from the presented pair is better?''}, and ours was preferred in 87.5 \% of cases.

\paragraph{Quality of conditional diffusion.}
We compare the quality of our conditional generation with TECA~\cite{zhang2023teca}.
We launch both of the methods for various prompts with several random seeds to obtain the hair volume that follows the desired text input.
The qualitative comparison can be seen in \Cref{fig:comparison_teca}.
While TECA produces great conditioning results most of the time, some severe artifacts are noticeable in the hair region.
Furthermore, the diversity of generations is limited, and we see some failure cases even for simple prompts like \textit{``A woman with straight long hair"}.
With our method HAAR, we provide a way to obtain detailed physically plausible geometry with large variations.

\subsection{Ablation study}
\vspace{-0.2cm}
\paragraph{Conditioning.}
The quality of the conditional diffusion model for hairstyle generation is highly dependent on the quality of the text encoder network $\tau({\cdot})$.
We ablate the performance of the conditional generation using pre-trained and frozen encoders, such as CLIP~\cite{clip}, BLIP~\cite{li2022blip} as well as a trained transformer network~\cite{transformer} implemented on top of a pre-trained BertTokenizer~\cite{BertTokenizer}.
For more details on the architecture, please refer to the supplemental material.
The intuition behind training additional networks for text encoding is that the quality of pre-trained encoders may be limited for a particular task (for example some specific hairstyle types), which results in wrong correlations between words and deteriorates the quality of the diffusion model.

\input{arxiv_figures/ablation_tab}

We evaluate the performance using semantic matching between text and generated 3D hairstyles.
Specifically, we use CLIP~\cite{clip} and compute the cosine distance between images and their respective text prompts.
To do that, we generate 100 hairstyles for 10 different prompts and then render from a frontal view using Blender~\cite{Blender}. 
\Cref{tab:ablation_csim} shows that the BLIP text encoder is providing the most effective conditioning.
To show the upper-bound quality of this metric ('reference'), we calculate the CSIM on our ground-truth dataset with prompts obtained via VQA.
\input{arxiv_figures/editing/imagic_interpolation}

\paragraph{Upsampling scheme.}
We ablate the performance of different upsampling schemes needed to obtain a full hairstyle from a set of guiding strands, which can be seen in \Cref{fig:upsampling_ablation}.
There is no one-to-one correspondence and during interpolation, a lot of artifacts can occur.
The most common artifact is a visible \textit{grid structure} which appears when using a Nearest Neighbour (NN) strategy.
Bilinear interpolation leads to \textit{scalp penetrations} due to averaging the nearest strands on top of the head, and it deteriorates the local shape of curls.
The computer graphics engines, such as Blender~\cite{Blender} and Maya~\cite{maya}, either do not provide enough control or require a lot of manual effort in setting up the optimal parameters for each hairstyle separately.
We find that the combination of NN and Bilinear using our proposed scheme leads to the best-looking results of renders.
Furthermore, adding noise in the latent space results in more realistic hairstyles.
Note, for visualization we show an example with a reduced density of around 15,000 strands; increasing it leads to less bald regions, especially, in the region of a partition.

\subsection{Hairstyle editing}

Similar to Imagic~\cite{kawar2023imagic}, we do text-based hairstyle editing, see \Cref{fig:imagic_interpolation_main}.
Given an input hairstyle and a target text that corresponds to the desired prompt, we edit the hairstyle in a way that it corresponds to the prompt while preserving the details of the input hairstyle.
To do that we first do textual inversion of the input hairstyle.
We obtain $e_{tgt}$ that corresponds to the target prompt $P$.
After optimizing it with a fixed diffusion model $\mathcal{D}_\theta$ using a reconstruction loss, we acquire $e_{opt}$. 
Conditioning on the obtained text embedding $e_{opt}$ does not lead to the same target hairstyle.
So, to provide a smooth transition, we freeze $e_{opt}$ and fine-tune  $\mathcal{D}_\theta$.
Finally, we linearly interpolate between $e_{tgt}$ and $e_{opt}$. 
For more information, please refer to the supplemental material.

\input{arxiv_figures/limitation/limitations}

\subsection{Limitations}

The quality of generated hairstyles is limited by the variety and quality of our dataset, in terms of both the diversity of geometry assets and the accuracy of textual annotations.
The main failure cases include the generation of hairstyles with scalp interpenetrations and lack of curliness for some extreme hairstyles, see \Cref{fig:limitations}.
In theory, these limitations can be addressed with a dataset that contains more diverse samples of curly hairstyles, as well as human-made annotations.
Especially, when used in a physics simulation, the interpenetrations can be resolved in a postprocessing step.
Another limitation of our method is that we only consider geometry, we do not generate the hair color and texture which would be an interesting direction for future work.

%% file: arxiv_figures/comparison/comparison_main.tex
\begin{figure*}
    \resizebox{\textwidth}{!}{
    \includegraphics{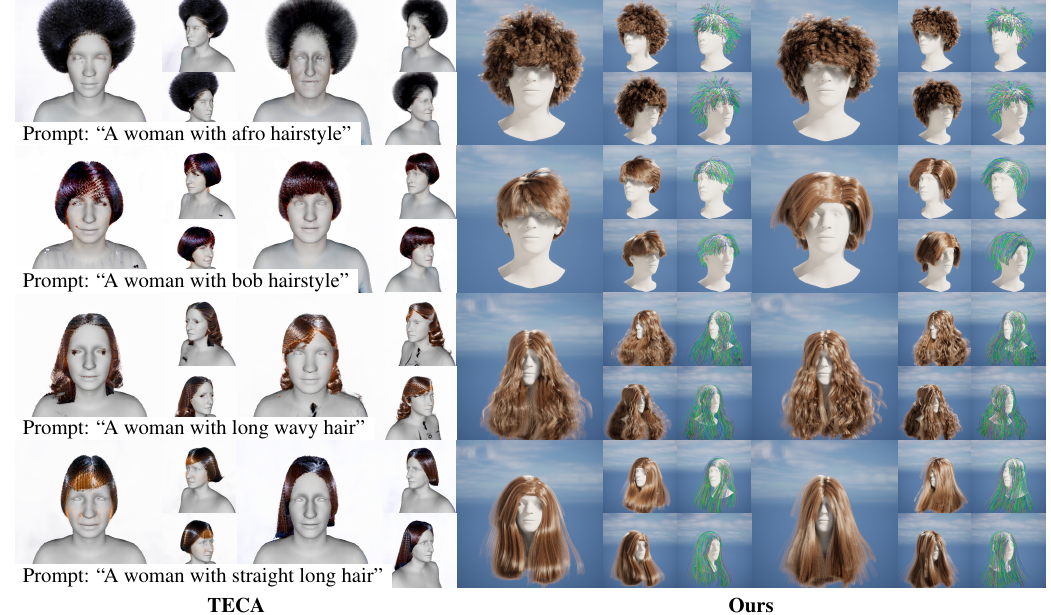}
    }
    \vspace{-0.4cm}
    \caption{ \textbf{Comparison.} Qualitative comparison of conditional generative models. We show several generations of TECA~\cite{zhang2023teca} and our model. For our results, we visualize the geometry obtained before (shown in pseudocolor) and after upsampling. Our model generates more diverse samples with higher-quality hairstyles. It is also worth noting that TECA, in some cases, does not follow the input descriptions well, producing short hair instead of long hair (bottom row). Digital zoom-in is recommended.}
    \label{fig:comparison_teca}

\end{figure*}

%% file: arxiv_figures/tab_uncond_comp.tex
\begin{table}
        \centering
        \begin{tabular}{ l|cccccc} 

            Method & \multicolumn{2}{c}{MMD$\downarrow$} & \multicolumn{2}{c}{COV$\uparrow$} & \multicolumn{2}{c}{1-NNA $\rightarrow$ 0.5} 
            \\
            \hline
            Neural Haircut~\cite{neuralhaircut}    & \multicolumn{2}{c}{$31507.7$} & \multicolumn{2}{c}{$0.18$} & \multicolumn{2}{c}{$0.34$}
            \\
            Our & \multicolumn{2}{c}{$21104.9$} & \multicolumn{2}{c}{$0.2$} & \multicolumn{2}{c}{$0.55$}\\
        \end{tabular}
        \caption{ \textbf{Comparison of unconditional diffusion models.} Our method generates samples with better quality and diversity.}
  \label{tab:uncond_comp}
\end{table}

%% file: arxiv_figures/upsample/upsample.tex
\begin{figure*}
    \resizebox{\textwidth}{!}{
    \includegraphics{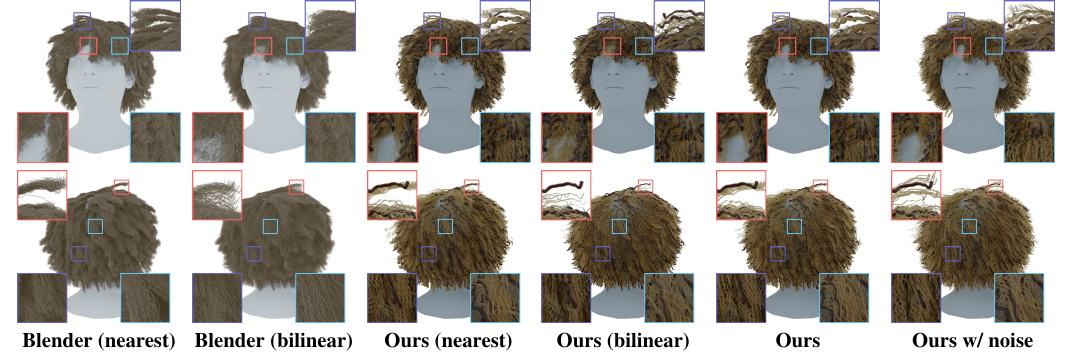}
    }
    \vspace{-0.4cm}
    \caption{\textbf{Upsampling.} Comparison of different upsampling schemes used to interpolate between guiding strands (shown in dark color). For visualization purposes here we show around 15,000 strands.  Blender interpolation is obtained in 3D space, while Ours is computed in latent space. Using the Nearest Neighbour in both variants produces better accuracy according to the guiding strand geometry (shown in dark color), but it results in an unrealistic global appearance. The bilinear schemes lead to the penetration of averaged hair strands and the loss of structure of the original guiding strands. Blending both these methods resolves proposed issues and results in realistic renders. Adding additional noise in latent space further increases realism and helps to get rid of the grid structure.}
    \label{fig:upsampling_ablation}
\end{figure*}

%% file: arxiv_figures/ablation_tab.tex
\begin{table}
        \centering
        \begin{tabular}{ l|cccc} 

            Text encoder & CLIP & BLIP & Transf. & Reference
            \\
            \hline
            CSIM  & 0.174 & 0.189  & 0.172 & 0.206
        \end{tabular}
        \caption{ \textbf{Conditioning.} Ablation on different conditioning schemes. With BLIP text encoder, we obtain better conditioning compared to CLIP and trainable Transformer network.}
  \label{tab:ablation_csim}
\end{table}

%% file: arxiv_figures/editing/imagic_interpolation.tex
\begin{figure*}
        \resizebox{\textwidth}{!}{
    \includegraphics{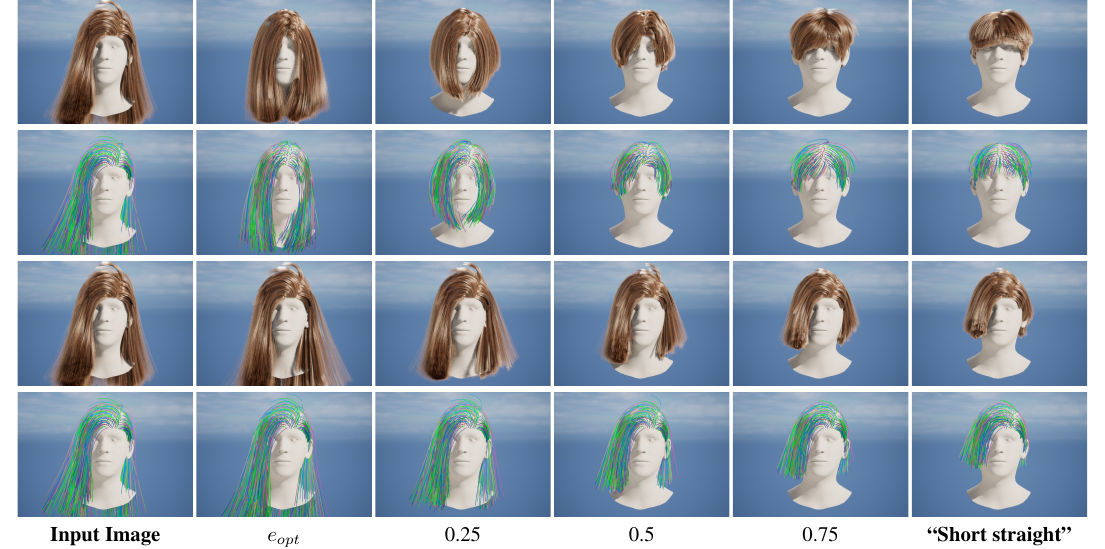}
    }
    \vspace{-0.3cm}
    \caption{\textbf{Hairstyle editing.} Similar to Imagic~\cite{kawar2023imagic}, we edit the input image using a text prompt. We provide editing results without additionally tuning the diffusion model (first two rows) and with it (second two rows). Finetuning the diffusion model results in smoother editing and better preservation of input hairstyle.}
    \label{fig:imagic_interpolation_main}
\end{figure*}

%% file: arxiv_figures/limitation/limitations.tex
\begin{figure}
    \begin{center}
    \includegraphics[trim=0 0 0 110,clip,width=0.23\textwidth]{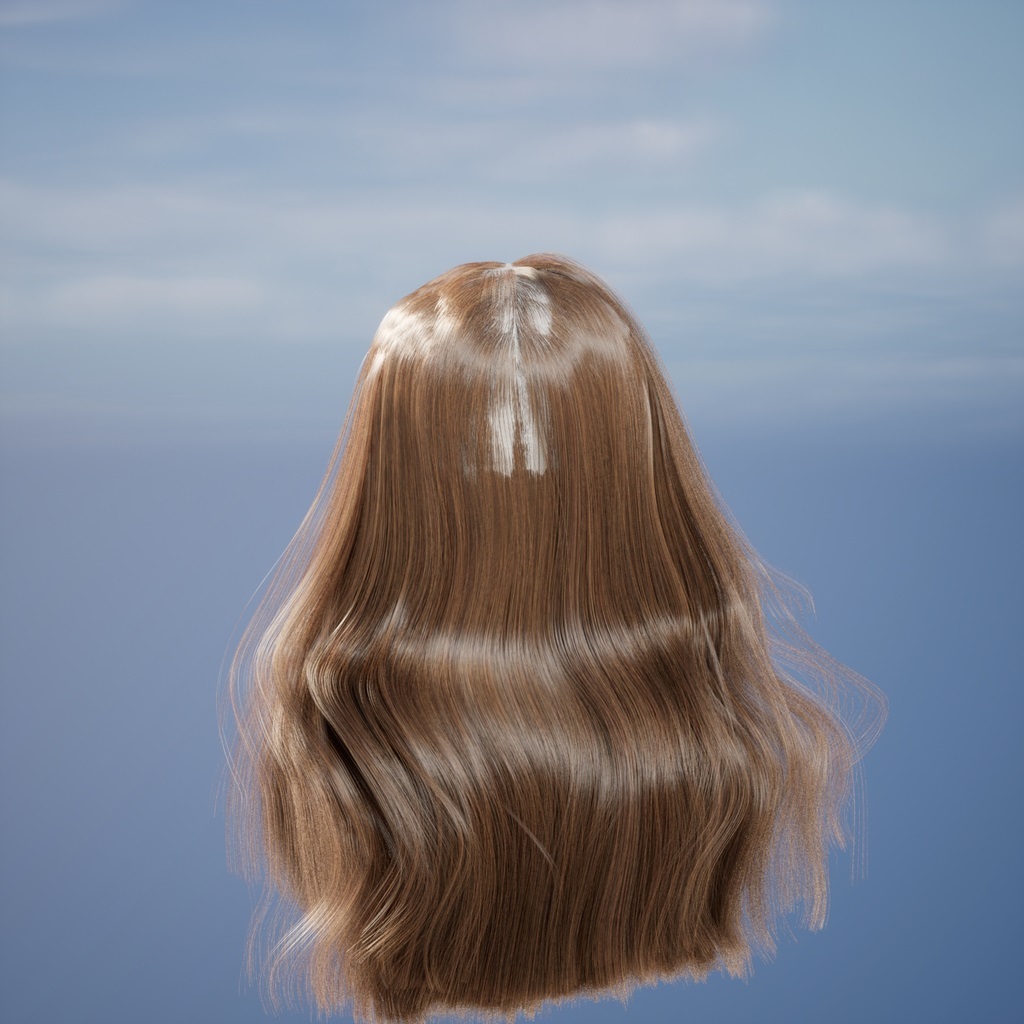}
    \includegraphics[trim=0 100 0 10,clip,width=0.23\textwidth]{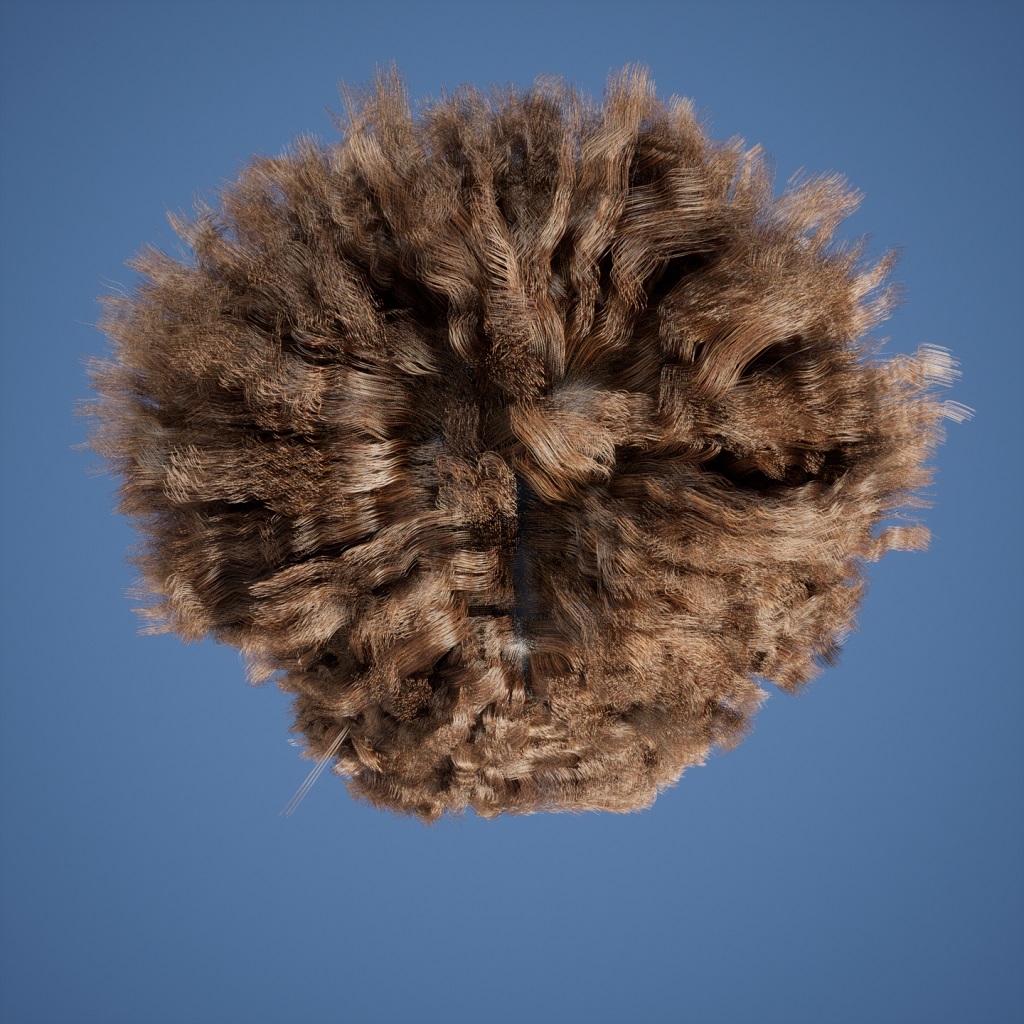}   \\    
    \end{center}
    \vspace{-0.3cm}
    \caption{\textbf{Limitations.} Our failure cases include penetration into the scalp region (left), which in principle can be resolved in a post-processing step. Additionally, for the afro hairstyles (right), the degree of strands' curliness needs to be increased.}\label{fig:limitations}
\end{figure}

%% file: arxiv_parts/conclusion.tex
\section{Conclusion}

We have presented HAAR, the first method that is able to conditionally generate realistic strand-based hairstyles using textual hairstyle descriptions as input.
Not only can such a system accelerate hairstyle creation in computer graphics engines, but it also bridges the gap between computer graphics and computer vision.
For computer graphics, generated hairstyles could be easily incorporated into tools like Blender for hair editing and physics-based animation.
For computer vision, our system can be used as a strong prior for the generation of avatars or to create synthetic training data of realistic hairstyles.
While being limited by data, we think that this method is a first step in the direction of controllable and automatic hairstyle generation.

%% file: arxiv_parts/acknowledgements.tex
\section*{Acknowledgements}
Vanessa Sklyarova was supported by the Max Planck ETH Center for Learning Systems. Egor Zakharov's work was funded by the ``AI-PERCEIVE'' ERC Consolidator Grant, 2021. We sincerely thank  Giorgio Becherini for rendering hairstyles and Joachim Tesch for realistic hair simulations.  Also, we thank Yao Feng and Balamurugan Thambiraja for their help during the project and Hao Zhang for aiding us with the TECA comparison. 

\paragraph{Disclosure.} MJB has received research gift funds from Adobe, Intel, Nvidia, Meta/Facebook, and Amazon.  MJB has financial interests in Amazon, Datagen Technologies, and Meshcapade GmbH.  While MJB is a consultant for Meshcapade, his research in this project was performed solely at, and funded solely by, the Max Planck Society.

%% file: arxiv_parts/method_suppmat.tex
\section{Implementation and training details}

\paragraph{Hairstyle diffusion model.} 
For conditional diffusion model, we use the U-Net architecture from~\cite{stable_diffusion} with the following parameters: $image\_size=32\times 32$, $input\_channels=64$, $num\_res\_blocks = 2$, $num\_heads = 8$, $attention\_resolutions = (4, 2, 1)$, $channel\_mult = (1, 2, 4, 4)$, $model\_channels = 320$, $use\_spatial\_transformer = True$, $context\_dim = 768$, $legacy = False$. 
Our training pipeline uses the EDM~\cite{Karras2022ElucidatingTD} library and  we optimize the loss function using AdamW~\cite{AdamW} with $learning\_rate=10^{-4}$,  $\beta=[0.95, 0.999]$, $\epsilon = 10^{-6}$, $batch\_size=8$, and $weight\_decay=10^{-3}$.

\paragraph{List of prompts.}
Below we include the list of prompts used during data annotation using a VQA model. After each of the prompts, we add \textit{`If you are not sure say it honestly. Do not imagine any contents that are not in the image. After the answer please clear your history.'} to the input.
\medskip
\begin{itemize}
    \footnotesize
    \item `Describe in detail the bang/fringe of depicted hairstyle including its directionality, texture, and coverage of face?'
    \item `What is the overall hairstyle depicted in the image?'
    \item `Does the depicted hairstyle longer than the shoulders or shorter than the shoulder?'
    \item `Does the depicted hairstyle have a short bang or long bang or no bang from frontal view?'
    \item `Does the hairstyle have a straight bang or Baby Bangs or Arched Bangs or Asymmetrical Bangs or Pin-Up Bangs or Choppy Bangs or curtain bang or side swept bang or no bang?'
    \item `Are there any afro features in the hairstyle or no afro features?'
    \item `Is the length of the hairstyle shorter than the middle of the neck or longer than the middle of the neck?'
    \item `What are the main geometry features of the depicted hairstyle?'
    \item `What is the overall shape of the depicted hairstyle?'
    \item `Is the hair short, medium, or long in terms of length?'
    \item `What is the type of depicted hairstyle?'
    \item `What is the length of hairstyle relative to the human body?'
    \item `Describe the texture and pattern of hair in the image.'
    \item `What is the texture of depicted hairstyle?'
    \item `Does the depicted hairstyle is straight or wavy or curly or kinky?'
    \item `Can you describe the overall flow and directionality of strands?'
    \item `Could you describe the bang of depicted hairstyle including its directionality and texture?'
    \item `Describe the main geometric features of the hairstyle depicted in the image.'
    \item `Is the length of a hairstyle buzz cut, pixie, ear length, chin length, neck length, shoulder length, armpit length or mid-back length?'
    \item `Describe actors with similar hairstyle type.'
    \item `Does the hairstyle cover any parts of the face? Write which exact parts.'
    \item `In what ways is this hairstyle a blend or combination of other popular hairstyles?'
    \item `Could you provide the closest types of hairstyles from which this one could be blended?'
    \item `How adaptable is this hairstyle for various occasions (casual, formal, athletic)?'
    \item `How is this hairstyle perceived in different social or professional settings?'
    \item `Are there historical figures who were iconic for wearing this hairstyle?'
    \item `Could you describe the partition of this hairstyle if it is visible?'
\end{itemize} 
\medskip

\paragraph{Text-based models.}
For the VQA, we found that ``LLaVA-v1.5'' model~\cite{liu2023improvedllava, liu2023llava} produces the best text descriptions with a relatively low hallucination rate. 
As shown in \Cref{tab:ablation_csim}, we experimented with different text encoder models.
We used ``ViT-L/14'' configuration for CLIP~\cite{clip} and ``blip\_feature\_extractor'' from~\cite{li-etal-2023-lavis} library for BLIP~\cite{li2022blip}.
In the ablation experiment, we compare its result with an optimizable Transformer~\cite{transformer} build on top of the pre-trained BERTTokenizer~\cite{BertTokenizer} from the transformers~\cite{wolf-etal-2020-transformers} library with configuration ``bert-base-uncased''.
For the Transformer network, we use BERTEmbedder from~\cite{stable_diffusion} with $n\_layer=6$, $max\_seq\_len=256$, $n\_embed=640$.

%% file: arxiv_parts/experiments_suppmat.tex
\section{Additional Ablations and Results}

\paragraph{Qualitative comparison.}
We show an extended comparison with TECA~\cite{zhang2023teca} with more complex prompts that show the compositional abilities of the models (see \Cref{fig:comparison_teca_suppmat}).
\input{arxiv_figures/comparison/comparison_suppmat}

\paragraph{Importance of classifier-free-guidance.}
To improve the sample quality of the conditional model, we use classifier-free-guidance~\cite{ho2021classifierfree}.
During training, we optimize conditional and unconditional models at the same time, by using text embedding with zeros in 10\% of cases.
During inference, we fix the random seed and show changes in sample quality, sweeping over the guidance strength $w$.
As we can see in \Cref{fig:cfg}, higher weights improve the strength of conditional prompts, but increasing it too much leads to out-of-distribution samples with a high degree of inter-head penetrations.
In our experiments, we fix the guidance weight to $w=1.5$.

\input{arxiv_figures/cfg/cfg}

\paragraph{Hairstyle interpolation.} 
We linearly interpolate between two text prompts $P_{1}$ and $P_{2}$ by conditioning the diffusion model $\mathcal{D}_\theta$ on a linear combination of text embeddings $ (1-\alpha)\tau (P_{1}) + \alpha\tau (P_{2})$, where $\alpha\in [0, 1]$, and $\tau$ is the text encoder.
For interpolation results obtained for different prompt pairs that differ in length and texture please see \Cref{fig:text_interpolation}.
One can notice that the interpolation between two types of textures, e.g. \textit{``wavy``} and \textit{``straight``} usually starts appearing for $\alpha$ close to 0.5, while length reduction takes many fewer interpolation steps.

\input{arxiv_figures/interpolation/interpolation}

\paragraph{Hairstyle editing.} 
For optimization $e_{opt}$, we do 1500 steps with the optimizer Adam with a learning rate of $10^{-3}$.
For diffusion fine-tuning, we do 600 steps with optimizer AdamW~\cite{AdamW} with a learning rate of $10^{-4}$,  $\beta=[0.95, 0.999]$, $\epsilon = 10^{-6}$, and weight decay $10^{-3}$.
Both stages are optimized using the same reconstruction loss used during the training of the main model.
The entire editing pipeline takes around six minutes on a single NVIDIA A100.
See \Cref{fig:imagic_interpolation} for more editing results with and without fine-tuning.
\input{arxiv_figures/editing/imagic_interpolation_suppmat}

\input{arxiv_figures/upsample/upsample_suppmat}

\paragraph{Upsampling scheme.} 
We provide more results on the different upsampling schemes for \textit{``long straight``} and \textit{``long wavy``} hairstyles (see \Cref{fig:upsampling_suppmat}).
While Blender~\cite{Blender} interpolation in 3D space produces either results with a high level of penetration (bilinear upsampling) or very structured (Nearest Neighbour) hairstyles, we are able to easily blend between two types in latent space, combining the best from the two schemes.
Adding noise helps eliminate the grid structure inherited from the nearest neighbor sampling and, thus, improves realism.
For noising the latent space, we calculate a standard deviation $\widetilde{Z}_{\sigma}\in\mathbb{R}^{1\times 1\times M}$ of latent map after interpolation $\widetilde{Z}\in\mathbb{R}^{N\times N \times M}$, where $N$ is a grid resolution and $M=64$ is the dimension of latent vector that encodes the entire hair strand. The final noised latent map is $\widetilde{Z}=\widetilde{Z}+ \widetilde{Z}_{\sigma}\odot X \odot Y$, where $X\in\mathbb{R}^{ N\times N\times 1}$ with elements $x_{ijk} \sim \mathcal{N}(0.15, 0.05)$, $Y \in\mathbb{R}^{ N\times N\times 1}$ with elements $y_{ijk} = 2q_{ijk} - 1, \quad \text{where} \quad q_{ijk} \sim \text{Bernoulli}(0.5)$.
In such a way, we independently add some small random noise to each latent vector on the mesh grid.

\paragraph{Generalization capabilities.}
Our conditional diffusion model can distinguish between different texture types, lengths, bangs, and some popular hairstyles, like the bob, and afro. 
It models the correlation between gender and hairstyle length, but at the same time, the capacity of the model is limited by the accuracy of the VQA and text encoder system.
Asking more general questions improves the generalization quality, but the answers may be less accurate and lead to additional noise during training.
To test the generalization capabilities of our model, we evaluate it on out-of-distribution prompts and attempt to generate hairstyles of particular celebrities.
We use ChatGPT~\cite{chatgpt} to describe the hairstyle type of a particular celebrity and use the resulting prompt for conditioning.
To our surprise, we find that even given the limited diversity of the hairstyles seen during training, our model can reproduce the general shape of the hairstyle.
We show results illustrating the generalization capabilities of our model by reconstructing celebrity hairstyles for \textit{``Cameron Diaz``} and \textit{``Tom Cruise``} (see \Cref{fig:generalization_suppmat}).
Between different random seeds hairstyles preserve the main features, like waviness and length, but could change the bang style. 
\input{arxiv_figures/generalization/generalization}

Finally, we show the results of our conditional model on different hairstyle types, by conditioning the model on hairstyle descriptions from~\cite{chatgpt} (see \Cref{fig:random_conditional_generation}). 
\input{arxiv_figures/conditional_generation/conditional_random_arxiv}

\paragraph{Simulations.} 
The hairstyles generated by our diffusion model are further interpolated to resolution $512\times512$ and then imported into the Unreal Engine~\cite{unrealengine} as a hair card.
We tested simulations in two scenarios: integration into a realistic game environment with manual character control as well as simple rotation movements for different types of hairstyles.
The realism of simulations highly depends on the physical characteristics of hair, e.g. friction, stiffness, damping, mass, elasticity, resistance, and collision detection inside the computer graphics engine.
An interesting research direction for future work may include the prediction of individual physical properties for each hairstyle that could further simplify the artists' work. 
For simulation results, please refer to the supplemental video.

%% file: arxiv_figures/comparison/comparison_suppmat.tex
\begin{figure*}
    \resizebox{\textwidth}{!}{
    \includegraphics{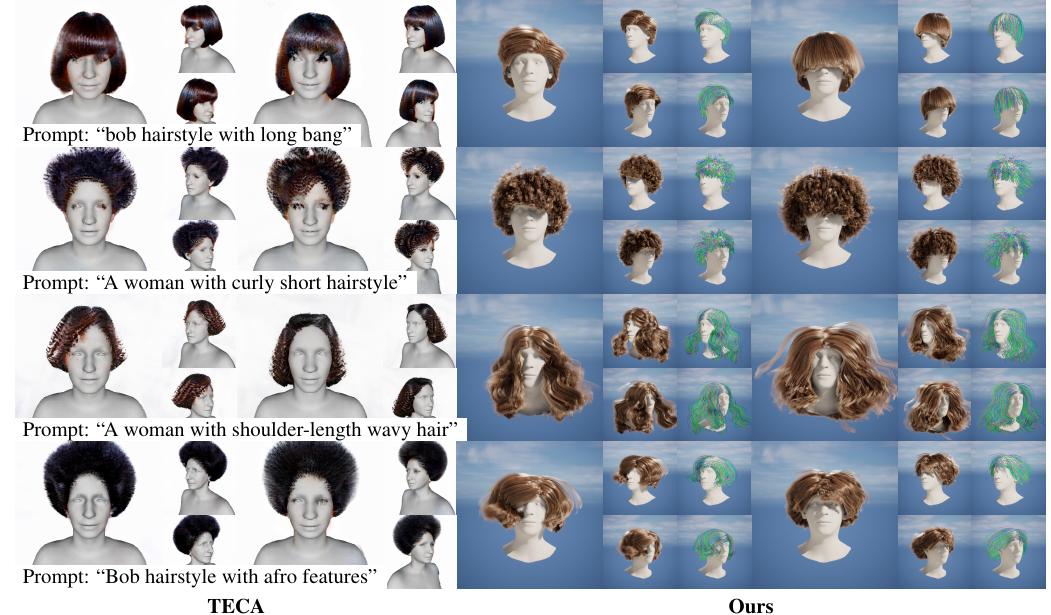}
    }
    \caption{ \textbf{Comparison.} Extended comparison with TECA. Our method produces higher quality samples with greater diversity than ones generated in TECA, and our representation allows the animation of the hair in a physics simulator.}
    \label{fig:comparison_teca_suppmat}
\end{figure*}

%% file: arxiv_figures/cfg/cfg.tex
\begin{figure*}
    \resizebox{\textwidth}{!}{
    \includegraphics{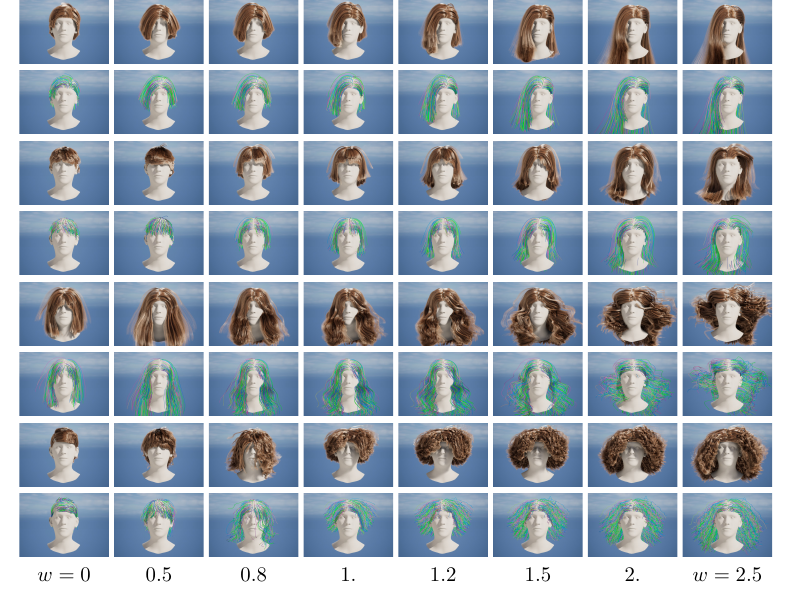}
    }
    \caption{\textbf{Classifier-free guidance.} Quality of samples during changing the guidance weight $w$ from 0 to 2.5. Weight $w=0$ corresponds to unconditional generation, while $w=1$ - to conditional. For $w>1$ we obtain over-conditioned results. In our experiments, we fix $w=1.5$, as higher weights lead to more penetrations and reduced realism. The first four rows correspond to generation samples for the prompt ``voluminous straight hair`` with two different random seeds, while the last four - for ``wavy long hair``.}
    \label{fig:cfg}
\end{figure*}

%% file: arxiv_figures/interpolation/interpolation.tex
\begin{figure*}
\centering
    \resizebox{1.\textwidth}{!}{
    \includegraphics{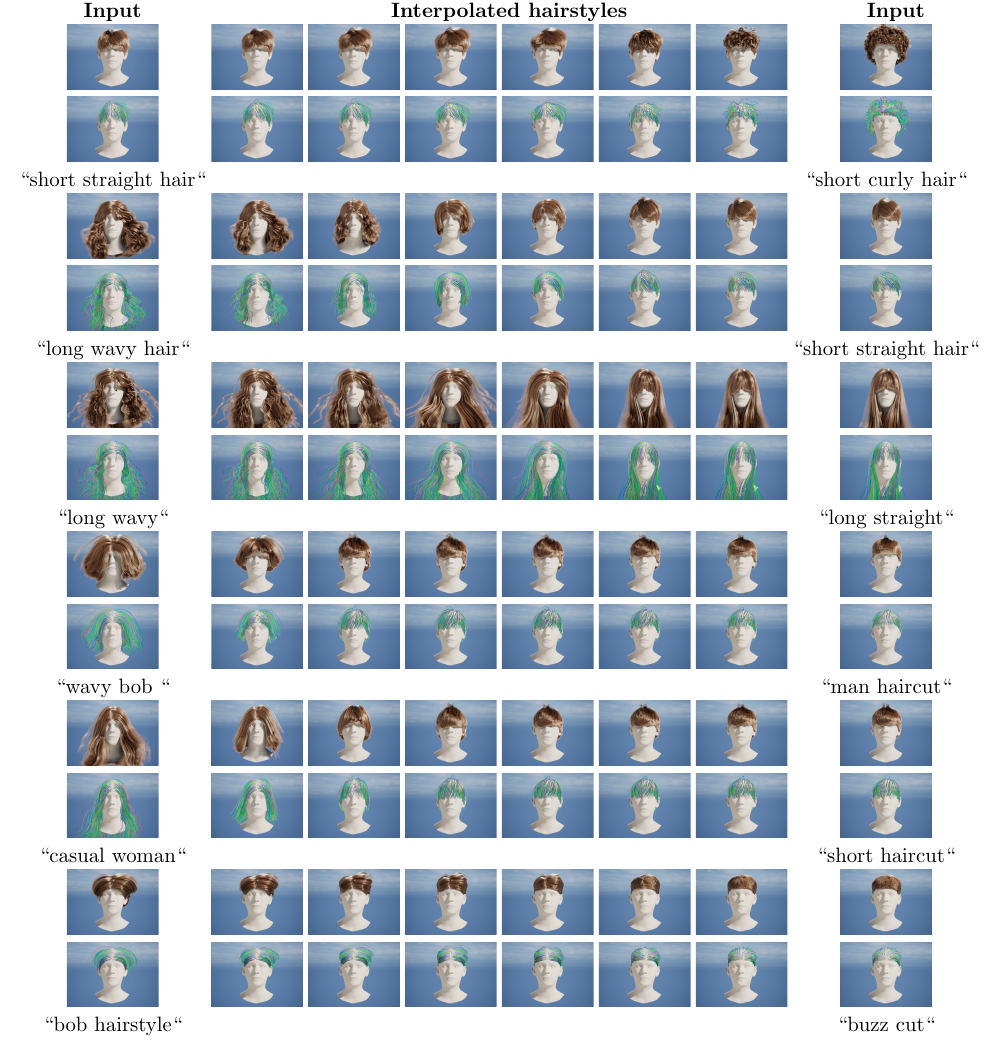}
    }
    \vspace{-0.3cm}
    \caption{\textbf{Hairstyle interpolation.} Linear interpolation between two given textual prompts. }
    \label{fig:text_interpolation}
\end{figure*}

%% file: arxiv_figures/editing/imagic_interpolation_suppmat.tex
\begin{figure*}
        \resizebox{\textwidth}{!}{
    \includegraphics{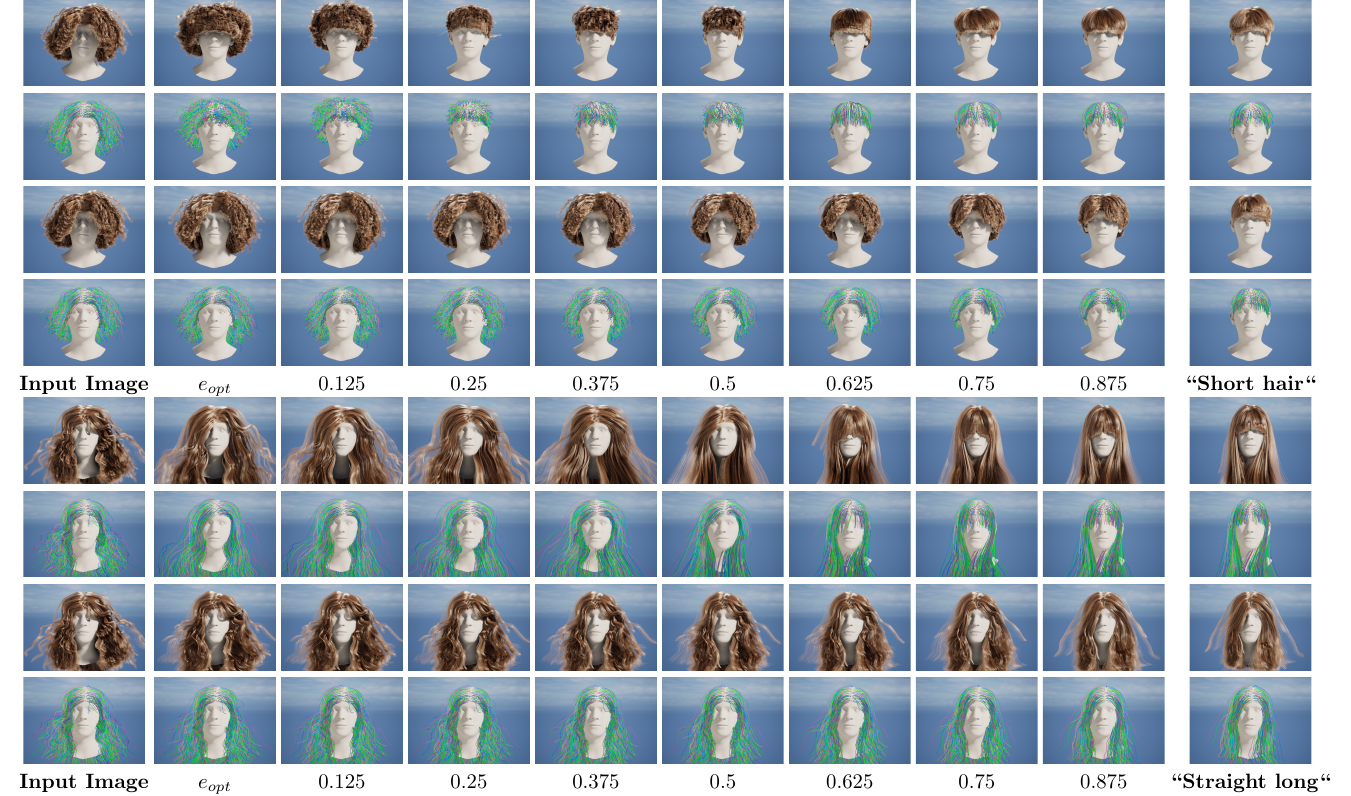}
    }
    \caption{\textbf{Hairstyle editing.} Extended editing results of our model. In each section of four images, we provide editing results without additionally tuning the diffusion model (first two rows) and with it (second two rows). Finetuning the diffusion model results in smoother editing and better preservation of input hairstyle.}
    \label{fig:imagic_interpolation}
\end{figure*}

%% file: arxiv_figures/upsample/upsample_suppmat.tex
\begin{figure*}
    \resizebox{\textwidth}{!}{
    \includegraphics{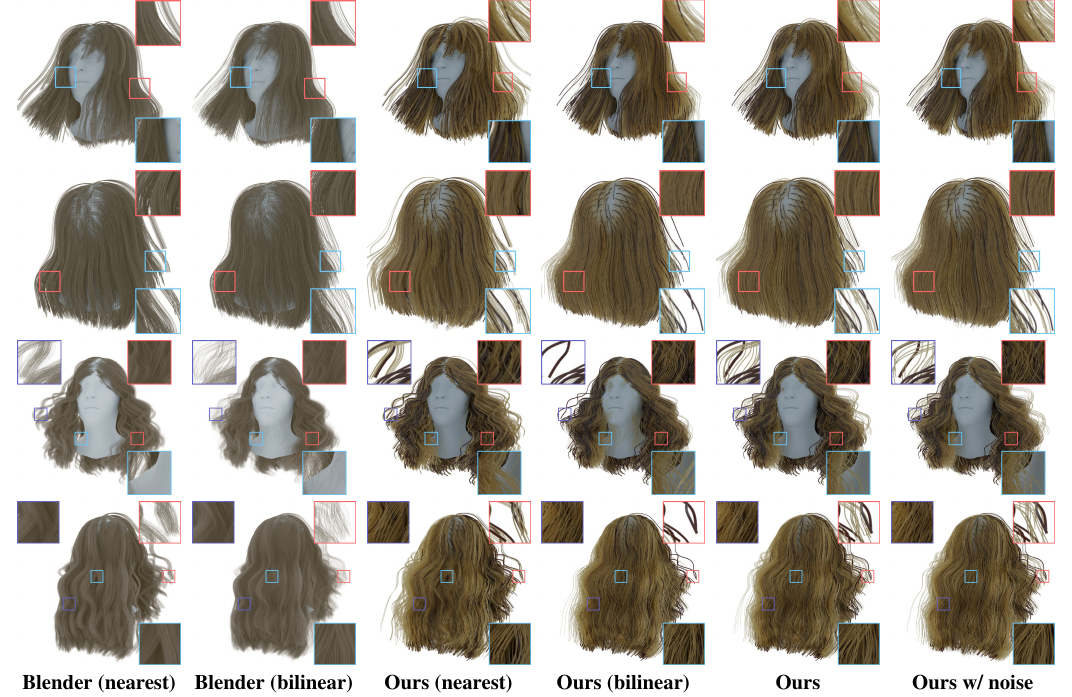}
    }
    \vspace{-0.2cm}
    \caption{\textbf{Upsampling.} Extended results on hairstyle interpolation between guiding strands obtained using different schemes. For better visual comparison, we interpolate hairstyles to around 15,000 strands and additionally visualize guiding strands (shown in dark color) for Ours methods with interpolation in latent space. Our final method with additional noise improves the realism of hairstyles by removing the grid-like artifacts.}
    \label{fig:upsampling_suppmat}
\end{figure*}

%% file: arxiv_figures/generalization/generalization.tex
\begin{figure*}
     \resizebox{\textwidth}{!}{\includegraphics{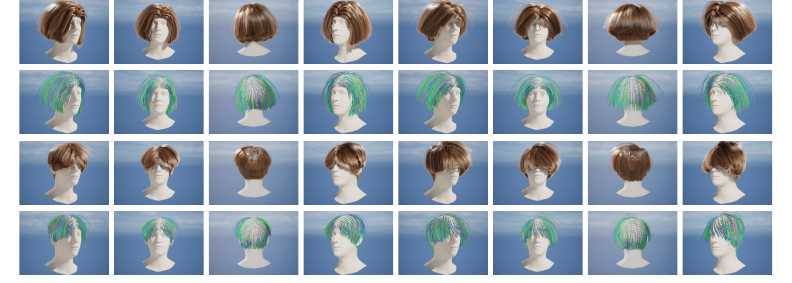}
     }
    \vspace{-0.2cm}
    \caption{\textbf{Generalization.} Hairstyles generated for celebrities  ``Cameron Diaz`` (first two rows) and ``Tom Cruise`` (last two rows) using descriptions from~\cite{chatgpt}. Several variations of hairstyles with corresponding guiding strands are generated for each celebrity.}
    \label{fig:generalization_suppmat}
\end{figure*}

%% file: arxiv_figures/conditional_generation/conditional_random_arxiv.tex
\begin{figure*}
    \resizebox{\textwidth}{!}{
    \includegraphics{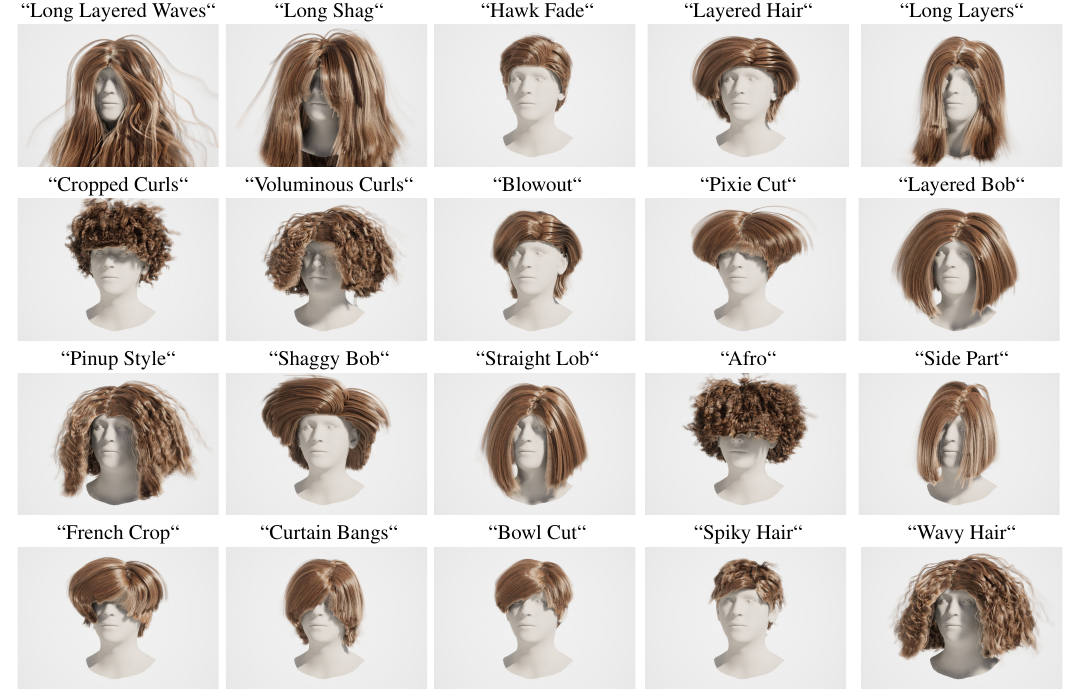}
    }
    \vspace{-0.2cm}
    \caption{\textbf{Conditional generation.} Random samples generated for input prompts with classifier-guidance weight $w=1.5$.}
    \label{fig:random_conditional_generation}
\end{figure*}